\documentclass[manuscript, journal, screen]{acmart}

\AtBeginDocument{%
  \providecommand\BibTeX{{%
    \normalfont B\kern-0.5em{\scshape i\kern-0.25em b}\kern-0.8em\TeX}}}

\setcopyright{acmlicensed}
\copyrightyear{2024}
\acmYear{2024}
\acmDOI{XXXXXXX.XXXXXXX}

\acmConference[]{}{}{}

\acmISBN{978-1-4503-XXXX-X/18/06}

\newcommand{\etal}{\textit{et al}.}
\newcommand{\etc}{\textit{etc}.}
\newcommand{\ie}{\textit{i}.\textit{e}.}
\newcommand{\eg}{\textit{e}.\textit{g}.}

\usepackage{amsmath}
\usepackage{stmaryrd}
\usepackage{mathtools}
\usepackage{acronym}
\usepackage{color}
\usepackage{xcolor}
\usepackage{verbatim}
\usepackage{booktabs}
\usepackage{siunitx}
\usepackage{suffix}
\usepackage{xstring}
\usepackage{xparse}
\usepackage{expl3}
\usepackage{mathrsfs}
\usepackage{tabularx}
\usepackage{multirow}
\usepackage{makecell}
\usepackage{array}
\usepackage[utf8]{inputenc} 
\usepackage{csquotes}

\usepackage{graphicx}
\usepackage{caption}
\usepackage{subcaption}
\usepackage{wrapfig}

\usepackage[T1]{fontenc}    
\usepackage{amsfonts}       
\usepackage{microtype}      

\AtBeginEnvironment{quote}{\singlespacing\small}
\usepackage{dirtytalk}
\usepackage{lscape}

\usepackage[ruled, linesnumbered]{algorithm2e}

\usepackage{subcaption}
\usepackage{wrapfig}
\definecolor{teally}{HTML}{51B3F9}

\usepackage{tikz}
\usetikzlibrary{arrows,backgrounds,calc}
\usepackage{relsize}
\usepackage{float}
\usepackage{kantlipsum} 
\usepackage{lipsum}
\usepackage{stfloats}

\begin{document}

\title[CARPs for Congestion Mitigation]{
Cooperative Advisory Residual Policies for Congestion Mitigation
}

\author{Aamir Hasan}
\email{aamirh2@illinois.edu}
\affiliation{%
  \institution{The University of Illinois Urbana-Champaign}
  \streetaddress{CSL Building, 
1308 W Main Street MC 228}
  \city{Urbana}
  \state{Illinois}
  \country{USA}}
  
\author{Neeloy Chakraborty}
\email{neeloyc2@illinois.edu}
\affiliation{%
  \institution{The University of Illinois Urbana-Champaign}
  \streetaddress{CSL Building, 
1308 W Main Street MC 228}
  \city{Urbana}
  \state{Illinois}
  \country{USA}}
  
\author{Haonan Chen}
\email{haonan2@illinois.edu}
\affiliation{%
  \institution{The University of Illinois Urbana-Champaign}
  \streetaddress{CSL Building, 
1308 W Main Street MC 228}
  \city{Urbana}
  \state{Illinois}
  \country{USA}}

\author{Jung-Hoon Cho}
\email{jhooncho@mit.edu}
\affiliation{%
  \institution{Massachusetts Institute of Technology}
  \city{Cambridge}
  \state{Massachusetts}
  \country{USA}}

\author{Cathy Wu}
\email{cathywu@mit.edu}
\affiliation{%
  \institution{Massachusetts Institute of Technology}
  \city{Cambridge}
  \state{Massachusetts}
  \country{USA}} 

\author{Katherine Driggs-Campbell}
\email{krdc@illinois.edu}
\affiliation{%
  \institution{The University of Illinois Urbana-Champaign}
  \streetaddress{CSL Building, 
1308 W Main Street MC 228}
  \city{Urbana}
  \state{Illinois}
  \country{USA}}

\renewcommand{\shortauthors}{Hasan \etal}



\begin{abstract}
Fleets of autonomous vehicles can mitigate traffic congestion through simple actions, thus improving many socioeconomic factors such as commute time and gas costs. 
However, these approaches are limited in practice as they assume precise control over autonomous vehicle fleets, incur extensive installation costs for a centralized sensor ecosystem, and also fail to account for uncertainty in driver behavior. 
To this end, we develop a class of learned residual policies that can be used in cooperative advisory systems and only require the use of a single vehicle with a human driver.
Our policies advise drivers to behave in ways that mitigate traffic congestion while accounting for diverse driver behaviors, particularly drivers' reactions to instructions, to provide an improved user experience. 
To realize such policies, we introduce an improved reward function that explicitly addresses congestion mitigation and driver attitudes to advice.
We show that our residual policies can be personalized by conditioning them on an inferred driver trait that is learned in an unsupervised manner with a variational autoencoder.
Our policies are trained in simulation with our novel instruction adherence driver model, and evaluated in simulation and through a user study (N=16) to capture the sentiments of human drivers.
Our results show that our approaches successfully mitigate congestion while adapting to different driver behaviors, with up to 20\% and 40\% improvement as measured by a combination metric of speed and deviations in speed across time over baselines in our simulation tests and user study, respectively.
Our user study further shows that our policies are human-compatible and personalize to drivers. 
\end{abstract}



\received{\today}
\received[revised]{}
\received[accepted]{}

\maketitle

\section{Introduction}
\label{sec:introduction}




Traffic congestion and its multitude of adverse effects are major burdens to modern society with nearly 75\% of US drivers perceiving congestion as a major problem in roadways~\cite{aaafts2017traffic}.
Congestion can cause stop-and-go waves~(the phenomenon where vehicles have to stop not soon after accelerating) that have lasting affects in a road network long after the initiation of the wave~\cite{stern2018dissipation}.
These waves, and congestion in general, increase commute times~\cite{epa}, emission rates~\cite{barth2008real}, fuel consumption~\cite{schrank2021urban}, and \textit{also} have a significant economic impact~\cite{goodwin2004economic}.
Therefore, mitigating congestion or even dampening stop-and-go waves would vastly improve various aspects of everyday life for people around the world~\cite{cambridge2004traffic}.

Solutions that address congestion mitigation exist on a spectrum between driver actuated solutions to completely autonomous solutions. 
Autonomous Vehicles (AVs) are arguably the most efficient solution to mitigate congestion~\cite{flow, stern2018dissipation, monache2019autonomous, monache2019feedback}.
Due to their near-perfect controllability, fleets of AVs or even one single AV can effectively smooth traffic flows~\cite{stern2018dissipation}.
However, current AV technology is inaccessible for deployment without incurring significant safety risks~\cite{kalra2014driving, dixit2016autonomous}.

\begin{figure}[b!]
    \centering
    \includegraphics[trim=0cm 1.3cm 0cm 1.2cm,clip=true,width=\textwidth]{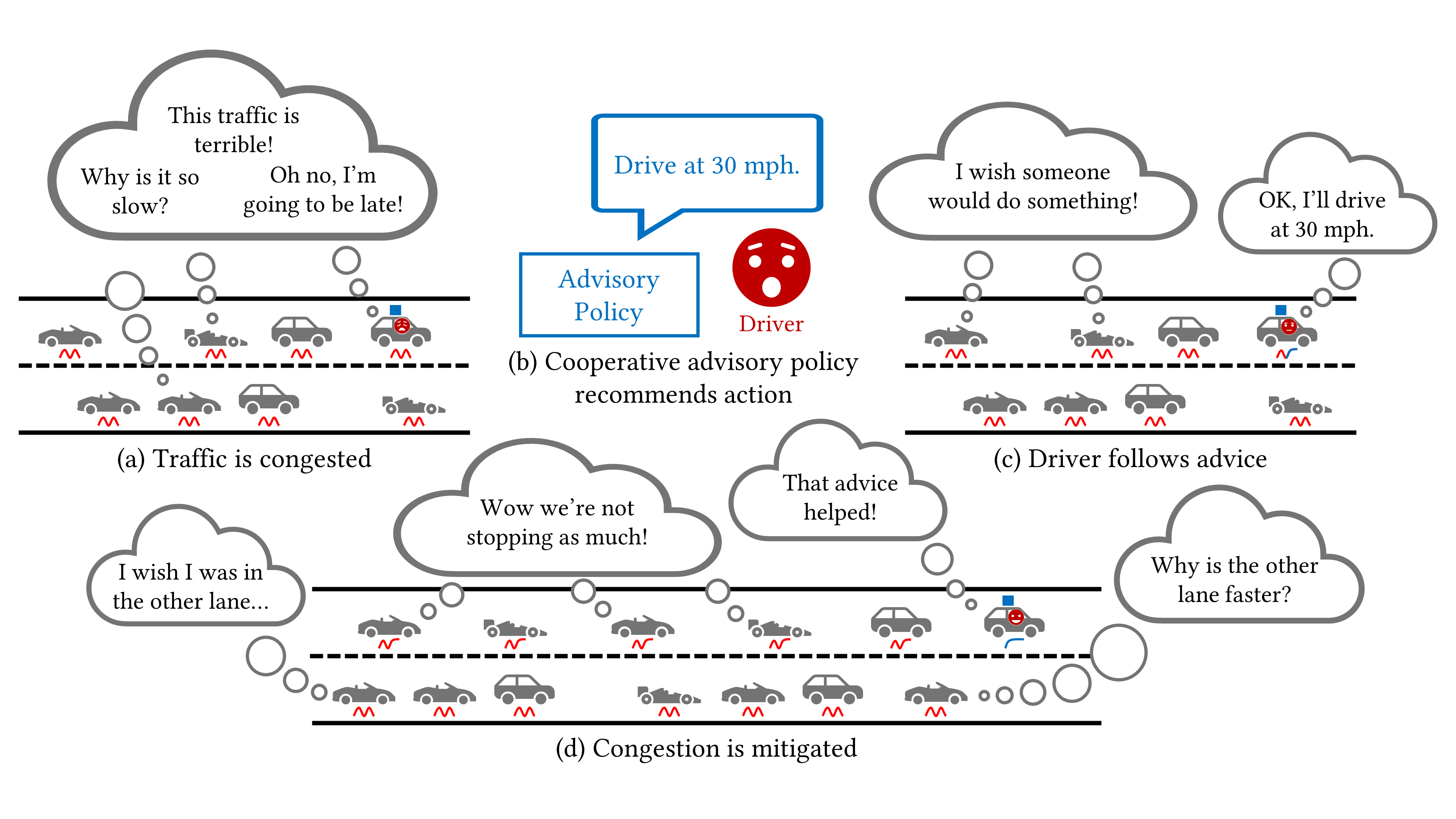}
    \caption{\textbf{An illustration of how Advisory policies can help mitigate congestion (Note: the lines under the cars represent their most recent speed profile and only the car with the driver has the advisory system.):} (a) Traffic congestion is present. (b) The Advisory Policy provides a suggestion to the driver of the vehicle with the blue square over it. (c) The driver follows the recommended advice. (d) The driver following the advice mitigates congestion.}
    \Description[Illustrations of advisory policies mitigating congestion]{An illustration showing different stages of how real-time advisory policies can aid in congestion mitigation. (a) Traffic on a two lane road with frustrated drivers. (b) The policy provides advice to one of the drivers. (c) The driver takes action according to the policies advice. (d) The driver has successfully mitigated congestion by following the advice of the policy in their lane. Drivers in the other lane are frustrated by continued congestion in their lane.}
    \label{fig:overview}
\end{figure}

Semi-autonomous solutions that utilize advanced driver assistance systems (ADAS) can act as substitutes to AVs to aid with congestion, \eg~Adaptive Cruise Control~\cite{hayat2023traffic, kim2022congestion}.
A significant asset of utilizing these solutions in mitigating congestion is their ability to alleviate the cognitive load placed on the driver~\cite{kummetha2020analysis}.
As such, multiple field experiments have been performed to prove their efficacy~\cite{stern2018dissipation, nice2023enabling}.
Although these systems can be effective, they incur a significant debt in deployment and are less reactive to drivers needs~\cite{hayat2023traffic, nice2023enabling}. 
Particularly, all field experiments conducted thus far require a central ecosystem that tracks vehicles in the network to determine what action the semi-autonomous solutions would need to take to be effective.
Drivers in these scenarios act as a supervisor and only step in during failure cases.

Alternately to semi-autonomous solutions, driver actuated solutions primarily rely on the driver and technology outside vehicles, as opposed to technology in(side) vehicles, to take an action to mitigate congestion. 
Examples of driver actuated solutions are speed limit signage and intersections~\cite{mirshahi2007active, nissan2011evaluation, stern2018dissipation}. 
Driver actuated solutions are \textbf{advisory systems} \ie~they advise drivers to behave in certain ways to mitigate congestion. 
Such solutions are primarily intended for encouraging safety and are hence provided en masse \ie~multiple drivers are given the same advice with the hope for a positive effect. 
Furthermore, while these solutions are helpful, they only address singular downstream effects of congestion. 
For example, intersections help control traffic waves, but increase carbon emission rates due to idling and slowdowns or speedups as vehicles approach the intersection~\cite{lertworawanich2021aco}.  
Speed limits are inefficient due to their static nature and are purposely mis-followed by drivers~\cite{mannering2009empirical, nissan2011evaluation}.
Therefore, a significant drawback of these solutions is their over-reliance on drivers -- which leads to failure as variations of driver behavior in following advice arise. 

Naturally, advisory systems that capitalize on AV technology and advise drivers individually have been proposed~\cite{pcp, flow, kreidieh2018dissipating, li2023stabilization, cho2023temporal}.
These learning-based advisory systems address drawbacks in both driver actuated solutions and autonomous solutions.
Particularly, they are more reactive to individual drivers rather than a collective by exploiting the human psyche~\cite{pcp, hasan2023perp}.
These advisory solutions have been popular due to advances in simulation and their exploration prowess~\cite{flow, kreidieh2018dissipating, cho2023temporal, hasan2023perp}.
 
In this work, we further the frontier of learning-based advisory systems by introducing a novel class of cooperative advisory residual policies.
We propose the utility of learned policies in combination with driver behavior models to enable adaption to differing driver behaviors. 
Our policies are trained in simulation with novel driver modeling that directly addresses how drivers follow  instructions in real-time. 
We incorporate driver behavior inference to enable our policies to react to the nuances in individual driving behaviors and hence cooperate with drivers and their preferences.
Furthermore, we directly incentivize our policies to provide advice that is easier to follow by human drivers through our novel reward function. 
While our policies are trained in simulation, our goal is to build a system that can be used by humans. 
Therefore, we conduct a driving simulator user study to evaluate our policies and compare them against baseline policies.
Through these various aspects, we address the drawbacks in driver actuated solutions while also utilizing the technological advances in autonomy. 
Figure \ref{fig:overview} shows an illustration of how such a policy can provide real-time advice to drivers to aid in mitigating congestion.


Our main contributions are:
\begin{enumerate}
    \item A novel class of cooperative advisory residual policies that provide real-time advice to drivers to mitigate congestion.
    \item A novel driver policy that models driver instruction following and hence aids in easier sim-to-real transfer.
    \item A simulation framework for driver-vehicle interactions in on-road settings that can work with a simulated driver and a human driver\footnote{Our framework and all code associated with the project will be released on acceptance}.
    \item A user study ($N=16$) that verifies the utility of advisory systems and our policies with quantitative and qualitative evaluations.
\end{enumerate}

To the best of our knowledge, this work is the first to build and test cooperative residual policies for congestion mitigation with user studies.
The rest of this paper is organized as follows: 
First, we discuss related work in Section \ref{sec:related-work}.
Then, we formally describe our problem statement and introduce preliminaries in Section \ref{sec:prelim}. 
Section \ref{sec:method} details our policies.
We then describe our simulator experiments and our user study in Section \ref{sec:experiments}.
Our results are discussed in Section \ref{sec:results}.
Finally, we conclude with a discussion on the limitations of our work and possible future directions in Sections \ref{sec:limitations} and \ref{sec:conclusion}.

\section{Related Works}
\label{sec:related-work}
In this section we provide an overview of the current landscape of congestion mitigation solutions in the form of model-based and learned strategies (see surveys and reviews:~\cite{papageorgiou2003review, shladover2005automated, cui2017stabilizing, afrin2020survey, kumar2023traffic}).
We also provide a background on advisory autonomy in transportation systems.


\subsection{Model-based Congestion Mitigation Strategies}
Model-based methods are primarily formulated as Adaptive Cruise Control (ACC) problems~\cite{davis2004effect, wang2016cooperative, monache2019feedback, stern2018dissipation}.
Particularly, fleets of vehicles controlled using variations of ACC and Cooperative ACC systems have been shown to smooth traffic flow in a road network~\cite{Rajamani2002semi, milanes2014cooperative, yu2021automated}.
These ACC approaches often attempt to optimize control decisions for alternative objectives (\eg~decreasing fuel consumption~\cite{van2006impact}),
use simplified modeling to extend to multi-agent system-wide objectives~\cite{kesting2007jam, au2014semi, monache2019feedback, martínez2023mitigation}, 
or apply more complex hierarchical formulations~\cite{fu2023cooperative} to lead to congestion mitigation.
While ACC technologies have matured since their inception, they are not without their drawbacks.
A primary drawback of this class of methods is the difficulty in practically realizing the extensive model design that is required for the rule-based controllers due to demanding connectivity requirements~\cite{milanes2014cooperative, nice2023enabling, monache2019autonomous}.
In contrast, our proposed methods rely on observations that can be captured with onboard sensors from \textit{a single vehicle} which significantly reduces the costs for adoption.
The efficacy of model-based solutions capitalizing solely on onboard sensors has been verified by \citet{nice2021can} in naturalistic studies with human drivers. 
However, model-based strategies do not adapt to driver preferences, or to the fast-paced changes in the road network caused by unpredictable disturbances. 
Learning-based strategies, including those presented in this article, aim to address these disadvantages through the use of data-driven methods and driver inclusive design.

\subsection{Learning-based Congestion Mitigation Strategies}
Learning-based, particularly Deep Reinforcement Learning (RL) based, approaches have been employed for congestion mitigation through infrastructure controls in the forms of variable speed limit signs, traffic lights, and ramps~\cite{li2017reinforcement, ke2021enhancing, yau2017survey}, and through developing longitudinal controllers for AVs~\cite{flow, kreidieh2018dissipating, vinitsky2018lagrangian, yan2021reinforcement, yan2022unified, han2024human}.
Due to their dependence on data, car following models such as IDM~\cite{idm}, versatile traffic simulators such as SUMO~\cite{sumo}, and learning frameworks such as Flow~\cite{flow} are paramount to the development of these methods.
This family of strategies introduce new reward functions and theoretical advances for learning policies~\cite{kreidieh2018dissipating, cho2023temporal} and aim to model a variety of traffic scenarios such as intersections~\cite{yan2021reinforcement, jayawardana2022learning}, and lane changes~\cite{flow, pcp}.
With this variance in techniques, RL-based controllers are successful in mitigating congestion.
However, most learning-based strategies are designed to be deployed with (connected) AVs and are thus currently impractical on real roads due to safety risks~\cite{naicAutonomousVehicles, verdictAV, forbesAV}, and thereby by-design infeasible for human use. 
In an effort to bridge this gap, \citet{pcp} propose Piecewise Constant (PC) policies to provide easy-to-follow advice to drivers through the employment of an action-extension period during which constant advice is provided to the driver.
A more in-depth discussion on PC policies is provided in Section \ref{sec:method-pcp}.
We build upon this piecewise constant ideology by augmenting PC policies with residual policies that address their drawbacks (expanded upon in Section \ref{sec:prelim-drawbacks}).
While PC policies and their adjacent classes of methods~\cite{li2023stabilization, cho2023temporal} are powerful in simulation, they lack empirical validation through user studies involving human participants, \textit{until now}.
In this article, we validate PC policies and our residual improvements through a user study.

\subsection{Cooperative Advisory Autonomy in Transportation Systems}

Advisory systems are popularly used in various intelligent transportation communities to improve safety and avoid the unintended side effects of human decision-making by providing instructions to the human drivers~\cite{kochenderfer2012next, panou2013railway, geertsma2023naval}.
For example, advisory systems have been extensively developed and deployed onboard aircrafts in the form of Airborne Collision Avoidance Systems (\eg~BCAS, TCAS, ACAS-X)~\cite{kochenderfer2012next, adkins1980air, orlando1983traffic}.
These airborne advisory systems provide pilots with instructions (\eg~climb or descend in altitude) to avoid collisions~\cite{kochenderfer2012next}.
Improvements in these systems have exploited technological advancements in onboard sensors (from beacons and sonars to GPS) and probability state estimation techniques using particle filters~\cite{kochenderfer2012next}. 
Additionally, these systems are designed to be robust (and hence cooperative) to pilot accuracy in following the advice through the use of Partially Observable Markov Decision Processes (PO-MDPs)~\cite{kochenderfer2012next}.

Advisory systems also exist in road transportation in the form of road signs and ADAS
~\cite{mirshahi2007active}.
Off-board strategies, such as speed limit signage (both constant~\cite{mannering2009empirical} and variable~\cite{nissan2011evaluation}), are unsuccessful as drivers tend to ignore them, and incur significant installation costs~\cite{stern2018dissipation}.
In contrast, advisory systems on board vehicles, through various ADAS (\eg~navigation assistants, collision detection systems), are more popular and relied upon by drivers. 
However, the onboard systems that are developed for congestion mitigation are model-based and pose the drawbacks discussed above.

The design and development of onboard advisory systems by combining the improvements in learning-based strategies discussed above, with the proved efficacy of advisory systems is promising.
In this work, we further develop advisory congestion mitigation strategies by improving driver modeling with a direct emphasis on advisory systems to improve simulations. 
We further develop trait inference modules that enable us to personalize our residual policies to different kinds of drivers that enable cooperative driver-friendly solutions. 


\section{Preliminaries}
\label{sec:prelim}

In this section we detail our problem statement and introduce the notation used in later sections.
We also describe the base policy for our residual policies, Piecewise Constant Policies (PCP)~\cite{pcp} and discuss their drawbacks to motivate our residual policies.

\subsection{Problem Definition}
\label{sec:method-prob-def}
Consider a road network with $N_V$ vehicles, where each vehicle $i \in \{1, \dots, N_V\}$ can be described using a state $s^i \in S$.
Let a single ego vehicle be driven by a human following some guidance, $a^{advice} \in A$.
The driver of the ego vehicle can be represented by the policy $\pi^{ego}: S \times A \rightarrow A$.
The other $N_V-1$ vehicles are driven by humans without any advice.
Each driver of the non-ego vehicles can be represented by a policy $\pi^i: S \rightarrow A$ that produces an action $a^{i} \sim \pi^{i}(s^i)~\forall i \in \{1, \dots, N_V-1\}$.
\[a^{ego} \sim \pi^{ego} (s, a^{advice}) \qquad a^{i} \sim \pi^{i} (s^i) \forall i \in \{1, \dots, N_V-1\}\]

We aim to find an Advisory Policy, $\pi^{advisor} : S \rightarrow A$, that provides the driver of the ego vehicle with an advised action $a^{advice}$ to mitigate congestion in the road network.
\[a^{advice} \sim \pi^{advisor}(s)\]

\textbf{Note:} In this work, we define all actions $a \in A$ as the speed of the vehicle. 
As the definitions above are agnostic to: 
(1) the type of action -- the action could be any other controllable attribute of the vehicle (\eg~acceleration, headway, \etc);
(2) the network configuration -- this formulation is general and can be extrapolated to any network with a defined state space $S$.
In general, our definition does not make assumptions regarding interactions between different vehicles in the network such as lane changes or fleet driving.
For the remainder of this article, we refer to the ego vehicle and its driver as the \emph{driver} or \emph{agent} interchangeably.

\subsection{Piecewise Constant Policies (PCP)}
\label{sec:method-pcp}

As mentioned in Section \ref{sec:related-work}, prior works have demonstrated that RL policies that are parameterized by the network configuration and produce a constant speed action can mitigate congestion~\cite{stern2018dissipation, flow}.
However, these policies are designed for AVs and not human drivers.
Therefore, they do not account for the intricacies in human behavior (\eg~reaction time).
\citet{pcp} aimed to address the above-mentioned issues by introducing PC policies. 

PC policies design every action output by the policy, $\pi^{PC}: S \rightarrow A$ to be held constant for a $\delta$ time period, called the hold-length or action extension period.
This piecewise constant nature is aimed to account for the time taken to transition between two consecutive advised actions, $a^{PC}_t$ and $a^{PC}_{t+\delta}$.
\[a^{advice}_t = a^{advice}_{t+1} = \dots = a^{advice}_{t + \delta - 1} = a^{PC}_t \sim \pi^{PC} (s_t)\]
We provide a brief introduction to the formulation of PC policies below but refer the reader to the works by \citet{pcp}, \citet{li2023stabilization}, and \citet{cho2023temporal} for more details, theoretical proofs and bounds regarding PC policies.

PC policies for providing instructions to drivers can be defined using an episodic Markov Decision Process (MDP).
An episodic MDP: $\mathcal{M}$, is defined as, $\mathcal{M}=(S, A, P, r, H, \delta, \gamma)$, where $S$ is the state space, $A$ is the action space, $P: S \times A \rightarrow S$ represents the transition probabilities, $r(a, s): S \times A \rightarrow \mathbb{R}$ is the reward function for a state-action pair $(s, a)$, $H$ represents the horizon, $\delta$ represents hold-length, and $\gamma \in [0, 1)$ is the discount factor.

A state, $s$, in the state space for PC policies, $S^{PC}$, is defined with regards to the ego vehicle as: 
$$s=\left(\frac{v_{ego}}{v_{max}}, \frac{v_{leader}}{v_{max}}, \frac{h_{leader}}{h_{max}}\right)$$
where $v_{ego}$ is the speed of the ego vehicle, $v_{leader}$ is the speed of the vehicle leading the ego vehicle, $h_{leader}$ is the headway distance between the ego vehicle and its leader, $v_{max}$ is the maximum speed allowed on the track, and $h_{max}$ is the maximum possible distance between two vehicles. 
We assume a fully observable state, $s \in S^{PC}$, for the agent.
The action space for PC policies is defined as $A^{PC}=\{0, \dots, A_{max}\}$, a discrete set of $\alpha$ equally spaced speeds in m/s, where $\alpha \text{ and } A_{max}$ are hyperparameters. 
The reward function, $R^{PC}$, used for the PC policies is the speed of the ego vehicle: $R^{PC} = v_{ego}$.
Note that the authors assume that the driver has exact control over the speed of the ego vehicle at all times \ie~ $\pi^{ego}(s, a^{advice}) = a^{advice}$.

In summary: At every $\delta$ time steps, the PC policy, $\pi^{PC}: S^{PC} \rightarrow A^{PC}$, samples an action $a^{PC}_t \in A^{PC}$ that represents the speed to be held for $\delta$ timesteps that maximizes the reward function, $R^{PC}$.

\subsection{Drawbacks of PC policies}
\label{sec:prelim-drawbacks}
With the inclusion of the hold-length, $\delta$, the above formulation of PC policies is important in defining human-compatible advisory systems.
However, this definition has three major drawbacks:

\begin{enumerate}
\item \textbf{Mis-specified Reward Function}: The reward function $R^{PC}$ described above only includes the speed of the ego vehicle. 
Such a formulation is not solely indicative of a congestion free network, where \textit{all vehicles} must travel at their maximum speed and not just the ego vehicle. 
Unfortunately, $R^{PC}$ encourages the policy to learn tailgating behavior unless the action space is restricted \ie~the $A_{max}$ parameter is set close to the ideal action as determined by the road network configuration.
We address this issue by specifying an additional reward term to minimize the headway between the vehicles.
We aim to incentivize the policy to mitigate congestion without constraining the action space with our improved reward function (see Section \ref{sec:method-reward}).

\item \textbf{Jerky actions}: 
The formulation also assumes that all changes between consecutive advised actions require the same amount of time to take effect \ie~the time taken by drivers to enact a change from $a^{PC}_t$ and $a^{PC}_{t+\delta}$ does not depend on $|a^{PC}_{t+\delta} - a^{PC}_{t}|$.
We show in Section \ref{sec:results-qualitative} through our user study that the advice provided by PC policies is not completely effective as it provides jerky actions that are difficult for drivers to follow.
To address this incompatibility, we provide further structure to the advised policy with our improved reward function. 

\item \textbf{Driver Modeling}:
While the inclusion of a hold-length aims at ensuring human-compatible policies, the formulation assumes that all drivers react to instructions similarly.
However, it is established that drivers exhibit unique behaviors while driving even though driving is a mostly structured task due to conventions and regulations
~\cite{brown2020taxonomy}.
We aim to account for the diverse driver behaviors by introducing a driver policy $\pi^{driver}$ that is detailed in Section \ref{sec:method-driver-traits}.
This enhanced driver behavior model is used in simulation during training for better sim-to-real transfer when testing with human drivers. 
Furthermore, we train a trait inference module (introduced in Section \ref{sec:method-dti}) to capture driver traits to personalize our policies. 
\end{enumerate}

In Section \ref{sec:results}, we show that these inclusions lead to an improvement in overall efficiency of the system and aid in congestion mitigation.
\section{Method}
\label{sec:method}

In this section we describe the different components of our system.
We address the drawbacks discussed in Section \ref{sec:prelim-drawbacks} by first detailing our proposed improved reward function. 
We then define our driver policy and the driver traits that are modeled by it.
Lastly, we introduce our class of Residual Policies (RP) and our Driver Trait Inference (DTI) module that is used to capture the traits of real drivers during evaluation.

\subsection{Reward Function for Human-Compatible Policies}
\label{sec:method-reward}
As mentioned in Section \ref{sec:prelim-drawbacks}, the reward function, $R^{PC}$ is mis-specified and hence leads to locally optimal policies that break in less constrained environments.
Therefore, we construct a novel reward function, $R^{RP}$, that aims at addressing the incomplete nature of $R^{PC}$ and synthesizing advised actions that are more accessible to human drivers.
Our improved reward function, $R^{RP}$ is defined as:
$$R^{RP}_t = \underbrace{\frac{1}{N_V}\sum_{i \in \{1, \dots, N_V\}}  \left[ \alpha_{speed}v^{i}_t - \alpha_{headway}h^i_t\right]}_{R^{CM}_t} \quad + \quad \overbrace{\alpha_{action}|a^{advice}_t - a^{advice}_{t - \delta}|}^{R^A_t}$$
where $\alpha_{speed}, \alpha_{headway}, $ and $\alpha_{action}$ are hyperparameters.

The reward term $R^{CM}$ ensures the dissipation of stop-and-go traffic waves.
The first term, $\alpha_{speed}v^{i}_t$ , aims at maximizing the speed of all vehicles.
Observe that this term is the only constituent term in $R^{PC}$.
The second term, $-\alpha_{headway}h^i_t$,  aims at minimizing the headway between all vehicles and their leaders. 
By incentivizing both behaviors in the reward function, we are able to achieve traffic flow where all cars are maximizing their speed while having a uniform minimum headway.
The inclusion of minimizing the headway uniformly for all cars ensures that there is no tailgating behavior, \ie~the ego vehicle is not the only car that is travelling at the maximum speed with the minimum headway.
While this formulation assumes full observability of all vehicles in the network, equivalent behavior can be obtained by minimizing the headway between the ego vehicle and its predecessor, and the headway between the ego vehicle and its follower.
This equivalence would only require observing the ego vehicle and the vehicles in its immediate vicinity.
We choose the former definition as the main contribution of our reformulation is not to create a perfect function but to address the misspecification of $R^{PC}$. 

The reward term, $R^{A}$, aims at encouraging the action advised by the policy to be more easily followable by human drivers.
The term rewards low deviations from the previous action and ensures that the change between consecutive action extension periods is minimized.
By reducing the deviation between consecutive actions, we intend to decrease the load placed on drivers \ie~drivers can react to changes in a more relaxed manner due to smaller, more gradual changes.
While the inclusion of $R^{A}$ aids in making the actions more human compatible, it does not address the difference in characteristics of different drivers.
We address these differences in driver characteristics and how to model them with our driver policy.

\subsection{Driver Policy and Driver Traits}
\label{sec:method-driver-traits}

The Intelligent Driver Model (IDM) has been extensively used to model realistic human driving~\cite{idm}. 
The model is successful at simulating different driving styles through its various hyperparameters~\cite{morton2017simultaneous, brown2020taxonomy}.
However, IDM is a \textit{car following model} and hence does not consider how human drivers follow and react to the instructions provided to them.
We address this gap by introducing a driver policy, $\pi^{driver}$, that enables easier sim-to-real transfer by directly modeling human instruction following. 

The driver policy, $\pi^{driver}: S \times A \rightarrow A$ , produces the action performed by the driver, $a^{driver}$, by using an input of the current state, $s \in S$ and the advised action, $a^{advice}$ provided by the Advisory policy, $\pi^{advisor}$. 
Note that the driver state space and the advisor state space are the same as both agents observe the same environment.

\begin{figure}[t!]
    \centering
    \includegraphics[width=\columnwidth]{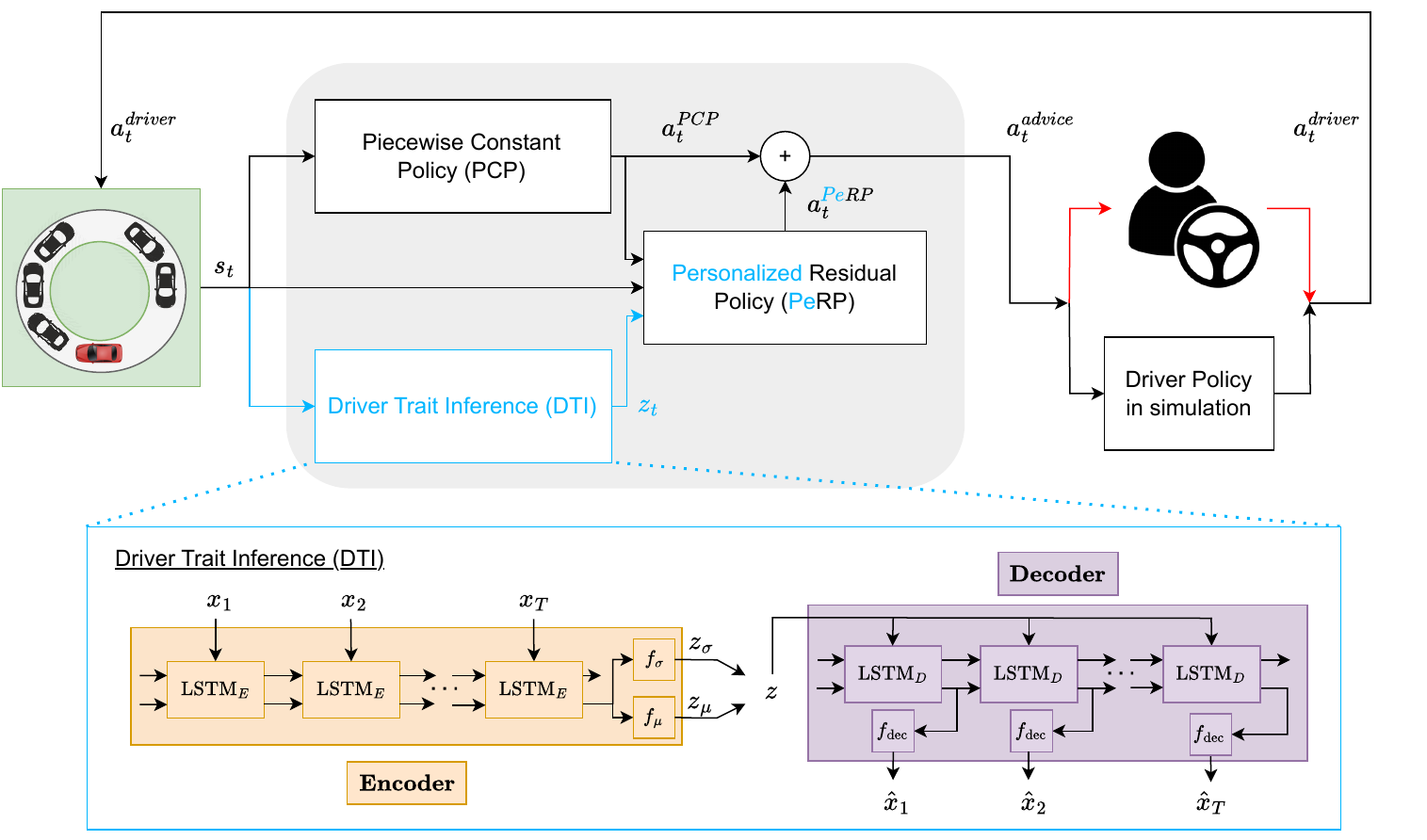}
    \caption{An overview of our method. Our residual policies offset the action output by the base policy to produce more actionable advice to drivers. Portions in \textcolor[HTML]{00B6FF}{blue} indicate the additional flow of data for our Personalized Residual Policy, PeRP, that also includes a latent vector depicting the driver's trait. Portions in \textcolor{red}{red} indicate the flow of data during evaluation with a human driver.}
    \Description[A flow diagram]{A flow diagram depicting the different components of the system.}
    \label{fig:method-overview}
\end{figure}

As shown in Figure \ref{fig:method-overview}, during simulation the action taken by the simulated driver is first pre-filtered by the driver policy. 
This filtered action, $a^{driver}$, is then passed into the SUMO simulator that internally uses IDM to change the simulated driver's action as specified. 
Through this process we are able to emulate diverse driver behaviors which we define to be the driver traits.
The driver traits parameterize $\pi^{driver}$ and hence inform \textit{how} drivers react to instructions.
Such an abstraction allows us to focus solely on the relationship between the driver and the advised action, while disregarding the mechanics involved in driving.
Similar methods have been used to simulate more realistic driving in simulation to aid learning with MDPs by \citet{janssen2019hidden} and \citet{jokinen2021modelling}.
%
We choose the following three traits to specify our driver policy:
\begin{enumerate}
    \item \textbf{Imperfect Instruction Following:} 
    Human drivers are incapable of holding the exact same action for a prolonged period of time due to the various hardware (in the vehicle) and physiological inconsistencies~\cite{ma2018relationship}. 
    Therefore, there are slight variations in the actions enacted by drivers. 
    We model this in our driver policy by the addition of standard normal Gaussian white noise. 
    \begin{gather*}
        \pi^{driver} (s, a^{advice}) = a^{advice} + k \qquad \text{ where } k \sim \mathcal{N}(0, 1)
    \end{gather*}

    \item \textbf{Reaction Time:} 
    Humans require a few seconds to be able to react to and perceive a change in action~\cite{droździel2020drivers, jurecki2017drivers}. 
    We model this in our driver policy as a time delay in action propagation. 
    Formally, this can be described as 
    \begin{gather*}
        \pi^{driver}_t(s, a^{advice}_t) = a^{advice}_{t-\sigma} \qquad        \text{ where } \sigma \sim U[\{2, 3, 4, 5, 6\}]
    \end{gather*}
    where $\sigma$ is the driver's reaction time. 
    We treat the initial period where $t < \sigma$ as a period with no advised action~\ie~the driver aims to drive safely (avoid collisions, stay on road \etc).
    We simulate $\sigma$ as a uniform random variable between 2 to 6 seconds~\cite{droździel2020drivers, jurecki2017drivers}. 
    
    \item \textbf{Intentional Instruction Offset:} 
    We also model intentional offsets from the advised actions, inspired by the intentional offsets that drivers exhibit with speed limits~\cite{mannering2009empirical}.
    Formally, we describe this as
    \begin{gather*}
        \pi^{driver}(s, a^{advice}) = a^{advice} + m \qquad \text{ where } m \sim U[\{-7.5, -5, -2.5, 0, 2.5, 5, 7.5\}]
    \end{gather*}
    where $m$ represents the speed that the driver intentionally deviates from the advised speed by.
    By modelling this intentional offset we also aim to reduce the sim-to-real and sim-to-sim gaps that might occur in our system (discussed in Section \ref{sec:results-qualitative}).
\end{enumerate}
During simulation, we combine all three traits to yield the driver policy. Formally, this policy can be described as follows:
\[
\pi^{driver}_t(s, a^{advice}_t) = a^{advice}_{t-\sigma} + m + k
\]
where $\sigma, m, $ and $k$ are as described above.
The inclusion of the driver policy enables us to train advisor policies that are more robust to usage by human drivers.

\newpage
\subsection{Residual Policies}
\label{sec:method-rp}
\begin{wrapfigure}[35]{r}{0.5\textwidth}
    \begin{minipage}{0.5\textwidth}
        \vspace{-0.4cm}
        \begin{algorithm}[H]
        \caption{The Residual Policy algorithm}
        \label{alg:rp}
        \DontPrintSemicolon
        \KwIn{Piecewise constant policy: $\pi^{PCP}$, Driver policy: $\pi^{driver}$, \textcolor{blue}{Driver Trait Inference model: $f_{trait}$}, Learning rate: $\lambda$, Hold Length: $\delta$, Warm-up time: $W$, Horizon: $H$, Observation length: $T$, Maximum training iterations: $n_{max}$}
        Initialize $\pi^{RP}_\theta$\;
        $iter \gets 0$\;
        \While{$iter \le n_{max}$ and not converged}{
            $s_t \gets$ Reset environment\;
            \For{$t \in \{1, ..., W\}$}{
                Sample $a_t \sim IDM(s_t)$\;
                Execute action $a_t$ in environment\;
                Observe $s_{t + 1}$\;
            }
            \While{$t \le W+H$}{
                Sample $a^{PCP}_t \sim \pi^{PCP}(s_t)$\;
                \textcolor{blue}{$z_t \gets f_{trait}({s_{t-T}, \dots, s_{t}})$\;}
                Sample $a^{RP}_t \sim \pi^{RP}_\theta(s_t, a^{PCP}_t\textcolor{blue}{, z_t})$\;
                $a^{advise}_t \gets a^{PCP}_t + a^{RP}_t$\;
                \For{$\Tilde{t} \in \{1, ..., \delta\}$}{
                    Sample $a_{t + \Tilde{t}} \sim \pi^{driver}(s_{t + \Tilde{t}}, a^{advised}_{t + \Tilde{t}})$\;
                    Execute action $a_{t + \Tilde{t}}$ in environment\;
                    Observe $s_{{t + \Tilde{t}} + 1}, r_{t + \Tilde{t}}$\;
                }
                $t \gets t + \delta$\;
            }
            Estimate loss
            $L_\theta(\theta)$ using $\{s_W, ..., s_{W+H}\}$, $\{a_W, ..., a_{W+H}\}$, and $\{r_W, ..., r_{W+H}\}$\;
            $\theta \gets \theta - \lambda \nabla_\theta L_\theta(\theta)$\;
            $iter \gets iter + 1$
        }
        \end{algorithm} 
    \end{minipage}
\end{wrapfigure}
Policies that are trained-from-scratch (TfS) to maximize the reward function $R^{RP}$ (Section \ref{sec:method-reward}) with the inclusion of the driver policy (Section \ref{sec:method-driver-traits}) could address the drawbacks of PC policies.
However, these TfS policies are inefficient to train primarily due to instabilities in early training stages that lead to agent collisions in simulation.
We therefore choose to train residual policies based on the PC policies and avoid these instabilities in training.
Residual policy learning also adds stability and reduces training time\footnote{The difference in training time between TfS policies and our residual policies is on the order of days for the same hyperparameters.}, which has led to its success in improving existing solutions in classic controllers~\cite{silver2018residual, johannik2019residual} and in predicting object dynamics~\cite{ajay2018augmenting}.
Additionally, by training a residual off of the PC policies we also retain their close-to-optimal performance in constrained scenarios.
While the trained PC policies could also be used to warm start the TfS policies, we choose to learn residual policies as they allow us to personalize the advised actions more easily.
This special class of residual policies, that we call Personalized Residual Policies (PeRP), are discussed in Section \ref{sec:method-perp}.

An overview of the flow of our policies are shown in Figure \ref{fig:method-overview}. 
The residual policies can be described using the same definition of an episodic MDP, $\mathcal{M}$, from Section \ref{sec:method-pcp} with the following modifications.
The state space, $s^{RP} \in S^{RP} = S^{PC} \times A^{PC}$, is a (3 + 1) dimensional tuple, with observed state space $S^{PC}$ and the action space, $A^{PC}$.
The residual action space is a discrete space of $n_{actions}$ speed offsets around the 0 offset, where the difference between contiguous elements is $\epsilon$, \ie~
$A^{RP} = \{-n_{actions} \epsilon, \dots, -\epsilon, 0, \epsilon, \dots, n_{actions} \epsilon\}$, where $\epsilon$ and $n_{actions}$ are a hyperparameters. 

At each timestep, $t$, the RP, $\pi^{RP}: S^{RP} \rightarrow A^{RP}$, samples a residual action $a^{RP}_t \in A^{RP}$.
Then the action to be held for $\delta$ timesteps to mitigate congestion is $a^{advice}_t = a^{PC}_t + a^{RP}_t$.
This advised action can be provided to the driver to reduce congestion and thereby improve emission rates in a cooperative manner. 

Algorithm \ref{alg:rp} details the training procedure for the residual policies.
Note that RP is also a PC policy as it assumes that actions will be held for $\delta$ timesteps.

\subsection{Driver Trait Inference (DTI)}
\label{sec:method-dti}
IDM driver trajectories have been used to infer driving styles with the help of Variational Autoencoders (VAE)~\cite{vae, morton2017simultaneous, liu2021learning}.
Inspired by these methods, we use an unsupervised VAE based model to infer the driver traits from driving trajectories \ie~we use driving trajectories to estimate the parameters (traits), of the driver policy, $\pi^{driver}$. 
Particularly, we aim to capture the reaction time and intentional instruction deviation traits discussed in Section \ref{sec:method-driver-traits}.
We provide a summary of our Driver Trait Inference model here but refer the reader to work to our previous work for a more detailed discussion~\cite{hasan2023perp}.

The DTI model uses a sequence of states of the agent, $\boldsymbol{x}=\{x_1, x_2, .., x_T\}$, where each $x_t \in S^{PC}$ and $T$ is the observation period, as input to predict a latent vector, $z \in \mathcal{L}$, to encode the agent's trait.
As the DTI method is completely unsupervised, any driver traits that manifest in the input trajectories can be captured \ie~as long as the trajectories generated due to the different driver traits result in separable data, the trait can be captured without any supervision.
For example, we used a similar model to capture a driver trait that was based on intentional instruction offsets~\cite{hasan2023perp}. 

Figure \ref{fig:method-overview} includes a schematic of our DTI model.
The DTI model consists of an LSTM encoder and an LSTM decoder. 
The encoder produces a latent vector $z$ from the observed trajectory $\boldsymbol{x}$ that can be used to indicate the drivers' traits. 
The decoder reconstructs the observed trajectory as $\boldsymbol{\hat{x}}$ from the latent vector $z$. 
During inference, only the encoder network is used to produce the latent vector $z$ that personalizes the action produced by the residual policy.
The model is trained with the loss function:
$$\mathcal{L}_{DTI} = \beta_{\text{recon}} \cdot {\left\| \boldsymbol{\hat{x}} - \boldsymbol{x} \right\|}_2 + \beta_{\text{KL}} \cdot D_{\text{KL}}(z_\mu,z_\sigma)$$ where $D_{\text{KL}}(\mu, \sigma)$ is the KL divergence between any Gaussian distribution $\mathcal{N}(\mu, \sigma)$ and the standard normal distribution $\mathcal{N}(0,1)$, and $\beta_{\text{recon}}$ and $\beta_{\text{KL}}$ are hyperparameters.

\subsection{Personalized Residual Policies}
\label{sec:method-perp}
The inference of the driver's traits allows us to learn personalized residual policies. 
We modify the state space of the residual policy to include the latent vector inferred by the DTI module. 
PeRP can be defined by the modified state space, $S^{PeRP} = S^{RP} \times \mathcal{L}$, where $\mathcal{L}$ is the latent vector space, and action space $A^{PeRP} = A^{RP}$.
In summary, at each timestep, $t$, the PeRP, $\pi^{PeRP}: S^{PeRP} \rightarrow A^{PeRP}$, samples a residual action $a^{PeRP}_t \in A^{PeRP}$ conditioned on the latent representation sampled from the DTI module, $z \in \mathcal{L}$. 
Note that this increases the dimension of the state  from $(3 + 1)$ to $(3 + 1 + |\mathcal{L}|)$, where $|\mathcal{L}|$ represents the cardinality of the latent space $\mathcal{L}$.
$$a^{PeRP}_t \sim \pi^{PeRP}(s_t, a^{PC}_t, z_t) \qquad \text{where}~s \in S^{PC} \text{, }  a^{PC} \in A^{PC} \text{, and } z_t \sim DTI(\boldsymbol{x}_t)$$
$$a^{advice}_t = a^{PC}_t + a^{PeRP}_t$$

Practically, Algorithm \ref{alg:rp} is modified for PeRP with the inclusion of the \textcolor{blue}{text shown in blue}.
Note that we assume that the driver trait does not change during the action extension period. We leave the relaxation of this assumption for future work.

While this formulation employs a single latent vector $z \in \mathcal{L}$ to represent all the driver traits, we can easily expand the definition of $\mathcal{L}$ to include multiple latent spaces.
For example, if $\mathcal{L}^A$ captures the latent vector for a driver trait, $trait^A$, and $\mathcal{L}^B$ captures the latent vector for a driver trait, $trait^B$, we can define $\mathcal{L} = \mathcal{L}^A \times \mathcal{L}^B$.
The latent input $z$ to PeRP would then be $z = (z^A, z^B)$ where $z^A \sim DTI^A(\boldsymbol{x})$ and $z^B \sim DTI^B(\boldsymbol{x})$.
Thus, enabling us to learn two or more latent representations to encode different traits or different combination of traits.
While such a formulation aides in using multiple combinations of traits, we note that increasing the number of latent representations could adversely affect the performance and training time of the policy.
Alternately, a single combined latent space could be used.
However, some driver traits might dominate over others in learning the combined latent vector.
Thus, catering for the inclusion of multiple latent vectors aids in providing uniform weights for all driver traits. 
Note, that this framework also allows in combining the latents in any weighted combination if so desired.

\section{Experiments}
\label{sec:experiments}
In this section we discuss our policies and the baselines used for comparison.
We also detail the environmental setup for the policies and summarize the components of our user study.

\subsection{The Policies}

We evaluate two residual policies:
\begin{enumerate}
    \item \textbf{Vanilla Residual Policy (RP)}: A residual policy that is trained with the new reward function $R^{RP}$ and our driver policy $\pi^{driver}$. This is the model described in Section \ref{sec:method-rp}.
    \item \textbf{Personalized Residual Policy (PeRP)}: A residual policy that includes a latent representation of the estimated driver trait in its state space in addition to the new reward function and the driver policy. 
    This is the model described in Section \ref{sec:method-perp}.
    We train PeRP with two separate DTI modules that capture latents for the reaction time and intentional instruction offset traits, respectively.
\end{enumerate}

We compare the performance of our policies with three baselines:
\begin{itemize}
    \item \textbf{Optimal Speed Limit (OSL)}: A policy that outputs a constant action that is calculated empirically and depends on the network parameters and is verified to maximize the average speed of all vehicles. 
    For our network, this action was a speed of $8.65 m/s$. 
    This policy is equivalent to one that is trained for $\delta=\infty$ and is analogous to a policy that always advises the speed limit.
    \item \textbf{PC policy (PCP)}: The base policy used to train all the residual policies as described in Section \ref{sec:method-pcp}.
    \item \textbf{Trait Aware Residual Policy (TA-RP)}: A residual policy that is provided with ground truth driver traits instead of an inferred trait. This policy represents ``oracle" performance.
\end{itemize}

Training details for all models used in this work are provided in Appendix \ref{sec:appendix-training}. 
We trained and evaluated all policy types for three hold-lengths, $\delta = \{50, 70, 100\}$.
We trained 5 policies for each combination of policy types and hold-length, and only the best performing policies were considered for evaluation.
The best performing PC policies were used as base policies for the residual policies.
To ensure uniform comparisons between the baseline and residual policies, the residual policies are only compared against the same base policy that they were trained with. 

\begin{figure}[t!]
    \centering
    \includegraphics[trim=0.2cm 0.1cm 0.2cm 2cm, clip=true, width=\textwidth]{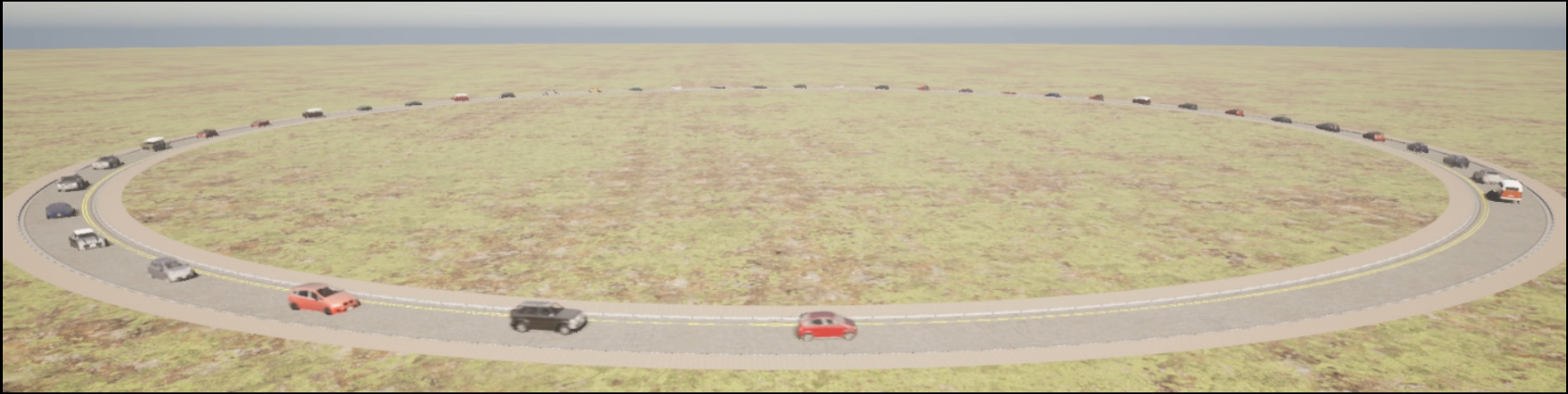}
    \caption{The simulation environment in CARLA. The \textcolor{red}{red} car in the center of the image is the ego vehicle. This image is inspired by Figure 1 in \citet{stern2018dissipation}.}
    \Description[circular ring road]{An image showing a circular ring road with one lane and 40 vehicles from the CARLA simulator.}
    \label{fig:environment}
\end{figure}

\subsection{Environment Setup}
The ring road network models an infinite highway and has been extensively used in congestion mitigation research~\cite{sugiyama2008traffic, tadaki2013phase, pcp, stern2018dissipation}.
We perform all operations on a single lane ring road with an inner circumference of $628m$ and outer circumference of $654m$ with $N_V=40$ vehicles.
We choose this configuration with a larger circumference (compared to other works) to ensure a comfortable driving experience (enough turning radius) for the participants in the user study.
We simulate the network in SUMO~\cite{sumo} and train our policies with the Flow framework~\cite{flow}.
For our user studies, we co-simulate the network and agents in the CARLA simulator~\cite{carla, sumo_cosim, hcd-workshop}.
Figure \ref{fig:environment} showcases our simulated environment in CARLA.
All non-ego vehicles were simulated using IDM, without the driver policy, during training and testing.

\begin{figure}[b!]
    \begin{minipage}{0.49\textwidth}
        \centering
        \includegraphics[trim=0cm 0.1cm 0cm 2cm, clip=true, width=\textwidth]{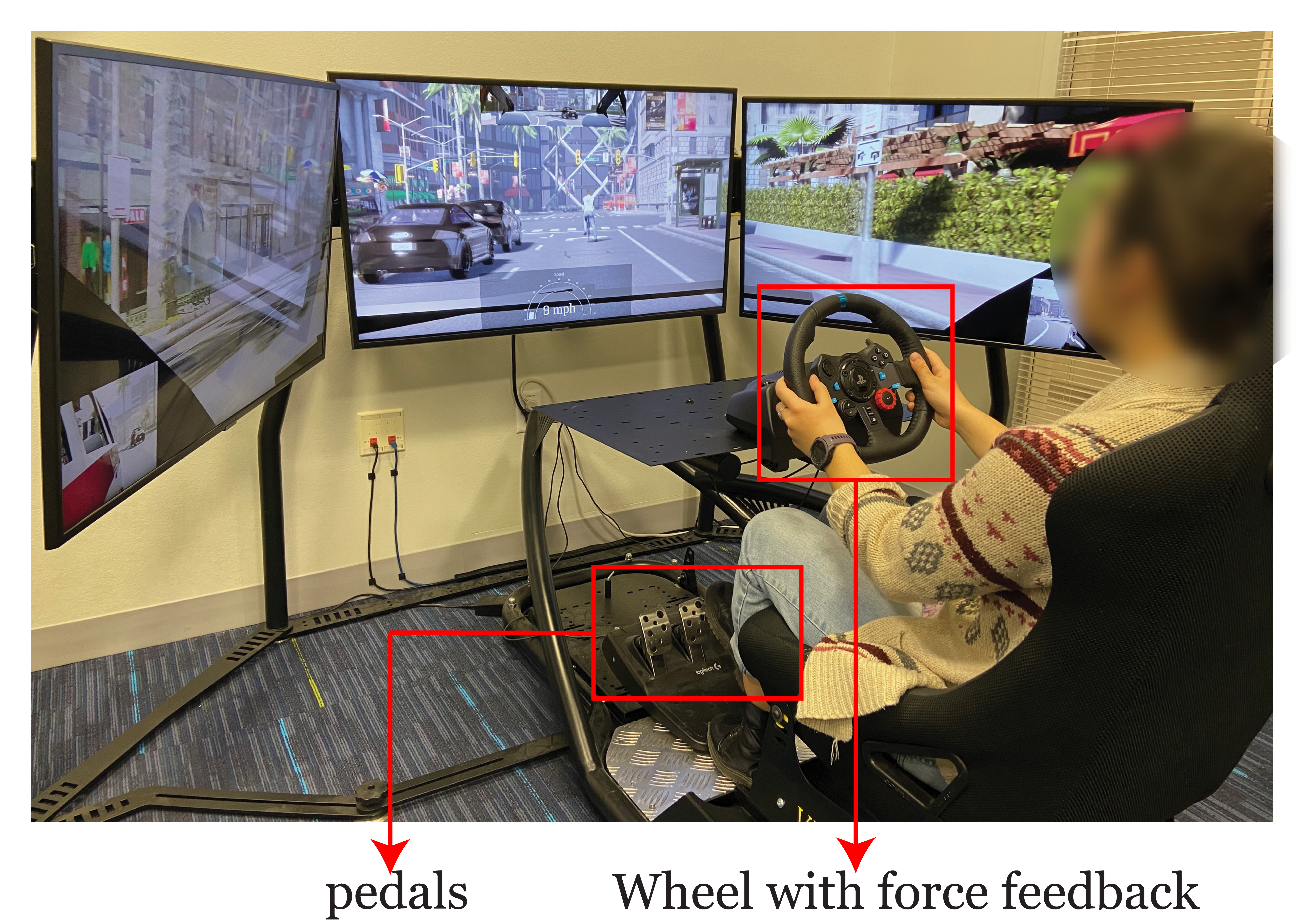}
        \caption{The driving simulator rig.}
        \Description[A driving simulator rig with three TVs]{An image showing a driving simulator rig with a wheel with force feedback and pedals to control the vehicle that is displayed on three TVs using the CARLA simulator.}
        \label{fig:driving-simulator}
    \end{minipage}\hfill
    \begin{minipage}{0.49\textwidth}
        \centering
        \includegraphics[trim=0cm 0.7cm 0cm 0cm,clip=true,width=0.8\textwidth]{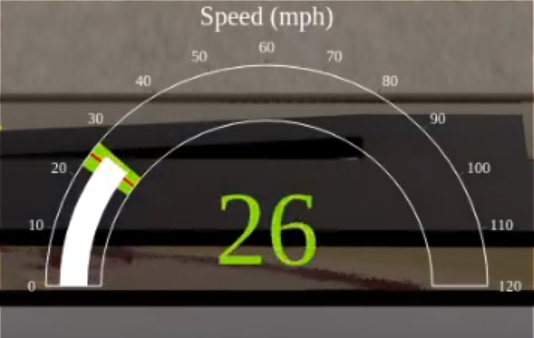}
        \caption{The speedometer showing the advised speed range (green area with red line) and the current speed. The current speed is displayed in green when the driver is within the advised range, and in white otherwise.}
        \Description[Speedometer with green range for advised speed]{An image showing a speedometer gauge with the speed displayed as a number in the center in large text. The gauge also has a green range with a red line indicating the advised speed to the driver.}
        \label{fig:ui}
    \end{minipage}
\end{figure}

\subsection{User Study}
In addition to testing the policies in simulation, we also test our system with a user study\footnote{The user study was approved by the IRB at the author's institution.}.
The goal of the user study was to assess the usability of the policies with human drivers.

\subsubsection{Participants}
The user study was conducted with N=16 participants (mean age=23; 10 males; 6 females).
All participants had at least a High School Diploma or equivalent and held a valid driving license in the US.
Participants reported average driving experience of $5.5 \pm 2.6$ years.
Additionally, 2 participants reported having some experience, 6 reported having little experience, and 8 reported having no experience with congestion mitigation policies. 
The study was conducted in a within-subjects manner where all participants experienced the same conditions (combination of $\delta$ and policy type) in a unique randomized order.
As PC policies have not been tested with human drivers, our user study also serves as a benchmark for their efficacy. 
We conducted a separate 3-person pilot study to validate our study procedure before recruitment for the larger study.

\subsubsection{Procedure}
Participants were first familiarized with the goal of the study and guided through an informed consent form.
Then, they were requested to fill a demographic and driving experience survey.
As a task summary, participants were told to drive as they would on the road \ie~avoid collisions and stay on the road, while trying their best to follow the instructions given by the policy.
Participants were explicitly told to ignore the advice of the policy if they determined that following the advice would lead to unsafe scenarios (collisions, off-roading~\etc).
Then, participants were provided with multiple practice runs\footnote{Only 2 participants required more than 1 practice run.} to get familiar with driving in the driving simulator and using the user interface (shown in Figures \ref{fig:driving-simulator} and \ref{fig:ui}, respectively).
Note that the policy used in the practice runs was not a policy that was evaluated to avoid bias (similar policies are discussed in Appendix \ref{sec:ablation-lower-amax}).
Participants could view a speedometer at the bottom of the simulator screen that showed them how fast they were driving in the form of a gauge as well as a large number (See Figure \ref{fig:ui}).
The advice given by the policy was displayed on the speedometer as a green range with a red line at its center. 
The red line indicated the exact speed advice, where the green range was a 1m/s area around the red line.
Participants were informed that they only had to maintain their speed within the green range as we had observed that following exact speeds would be hard during our system tests and pilot study. 

In total, 9 trials were conducted with each trial lasting at most 5 minutes, with three 5-minute breaks every 3 trials. 
After each trial, the participants completed a post-trial questionnaire that included: (1) the NASA Task Load Index (TLX) Questionnaire~\cite{hart1986nasa}; (2) the System Usability Scale (SUS) Questionnaire~\cite{brooke1996quick}; (3) a 5-point Likert scale question on how likely they were to use the policy in the real world. 
Participants were also asked for subjective feedback regarding their experience after each trial.
Note that participants were unaware of which policy they were testing in each trial.
All questionnaires and study procedures can be viewed on the project webpage\footnotemark[1].

\subsection{Metrics}
\subsubsection{Objective Metrics}
Given $\boldsymbol{v}$ is a collection of the speeds of the ego vehicle during the trial,
the policies were evaluated with following objective metrics:
\begin{itemize}
    \item \textbf{Average Speed ($\mu$):} The average speed of ego vehicle. 
    A higher value indicates good performance as the vehicle would travel as fast as possible in the absence of congestion.
    $$\mu(\boldsymbol{v}) = \frac{1}{T}\sum_{t=\{1, \dots, T\}}{v(t)}$$
    \item \textbf{Standard Deviation in Speed ($\sigma$):} The standard deviation in speed of the ego vehicle. 
    A lower value indicates good performance as more consistent speeds imply an absence of congestion. 
    $$\sigma(\boldsymbol{v}) = \sqrt{\frac{1}{T}\sum_{t=\{1, \dots, T\}}\left({v(t) - \mu(\boldsymbol{v})}\right)^2 }$$
    \item \textbf{Congestion factor (CF):} Congestion is defined to be mitigated when the vehicle is \textit{consistently} driving fast ~\cite{stern2018dissipation}. 
    Therefore, we require a metric that is a combination of the speed and the standard deviation in speed to properly compare different policies.
    We introduce the Congestion Factor (CF) metric to be defined as a combination of the above two metrics as:
    $$CF(\boldsymbol{v}) = \mu(\boldsymbol{v}) - \log_{10}{\sigma(\boldsymbol{v})}$$
    A higher value indicates better performance. 
    This metric rewards high speeds and low standard deviations in speeds.
    Deviations of less than $1 m/s$ are rewarded at an exponential rate as they indicate more mitigation which is harder to achieve based on empirical evidence.
\end{itemize}

\subsubsection{Subjective Metrics}
We evaluate the user's experience with the policies with the NASA Task Load Index Questionnaire (NASA TLX)~\cite{hart1986nasa} and the System Usability Scale (SUS) ~\cite{brooke1996quick}. 
Specifically, we use the average of all scores in the NASA TLX Questionnaire, also known as the Raw TLX Score, and the generic SUS Score as defined by ~\citet{brooke1996quick}. 
The NASA TLX questionnaire captures various attributes that indicate the mental workload due to a task, whereas the SUS questionnaire captures various attributes that include the usability of the system for a particular task.
A low Raw TLX score, and high SUS score indicate better performance.
The use of the questionnaires helps us measure how user friendly the policies can be.
The use of the 5-point Likert scale question about real-world use of the policy aided in directly querying the usability of the policies.
Short semi-structured interviews were conducted at the end of each trial to collect subjective feedback regarding the participants driving experience with the policy.

\section{Results and Discussion}
\label{sec:results}
In this section, we discuss the results of our experiments in simulation and through the user study. 
Analysis of the Driver Trait Inference module(s) is presented in Appendix \ref{sec:appendix-dti}.
All analysis presented here is with respect to CF as it provides a combination score of all other metrics to determine the level of congestion in the network.

\begin{table*}[b!]
    \centering
    \caption{Results comparing the performance of our policies with the baselines in simulation for 100 iterations. \textbf{Note:} $^{\dag}$~results are independent of $\delta$.}
    \resizebox{\textwidth}{!}{
    \begin{tabular}{l  c c c   c c c   c c c}
        \toprule
         \multirow{2}{*}{\shortstack[l]{Policy \\ Type}} & 
         \multicolumn{3}{c}{$\delta=50~(5\text{s})$} & 
         \multicolumn{3}{c}{$\delta=70~(7\text{s})$} &
         \multicolumn{3}{c}{$\delta=100~(10\text{s})$} \\
         \cmidrule(lr){2-4} \cmidrule(lr){5-7} \cmidrule(lr){8-10} 
         & CF $(\uparrow)$ &$\mu(\uparrow)$& $\sigma(\downarrow)$
         & CF $(\uparrow)$ &$\mu(\uparrow)$& $\sigma(\downarrow)$
         & CF $(\uparrow)$ &$\mu(\uparrow)$& $\sigma(\downarrow)$\\
         \midrule
         OSL$^\dag$& 
         6.50 $\pm$ 1.98 & 6.54 $\pm$ 2.21 & 1.59 $\pm$ 1.20 &
         6.50 $\pm$ 1.98 & 6.54 $\pm$ 2.21 & 1.59 $\pm$ 1.20 &
         6.50 $\pm$ 1.98 & 6.54 $\pm$ 2.21 & 1.59 $\pm$ 1.20 \\
         PCP&  
         6.83 $\pm$ 2.28 & 7.29 $\pm$ 1.33  & \textbf{3.06 $\pm$ 1.02} & 
         6.56 $\pm$ 1.72 & 6.98 $\pm$ 1.76 & \textbf{2.25 $\pm$ 1.15} &
         6.00 $\pm$ 2.26 & 6.39 $\pm$ 2.32 & \textbf{2.36 $\pm$ 1.34} \\ 
         TA-RP& 
         7.34 $\pm$ 0.56 & 7.79 $\pm$ 0.56 & 3.09 $\pm$ 0.79 & 
         6.63 $\pm$ 1.28 & 7.05 $\pm$ 1.23 & 2.69 $\pm$ 0.81 &
         7.36 $\pm$ 0.64  & 7.72 $\pm$ 0.71 & 2.67 $\pm$ 1.13 \\ 
         RP&
         \textbf{7.42 $\pm$ 0.34} & \textbf{7.86 $\pm$ 0.31} & 3.26 $\pm$ 0.85  & 
         \textbf{7.31 $\pm$ 0.59} & \textbf{7.71 $\pm$ 0.62} & 2.80 $\pm$ 0.93 &
         7.27 $\pm$ 0.73 & 7.62 $\pm$ 0.83 & 2.75 $\pm$ 1.33 \\ 
         PeRP& 
         7.27 $\pm$ 0.66 & 7.70 $\pm$ 0.65  & 3.10 $\pm$ 0.91 & 
         7.15 $\pm$ 0.91 & 7.55 $\pm$ 0.95 & 2.83 $\pm$ 1.07 &
         \textbf{7.36 $\pm$ 0.44} & \textbf{7.80 $\pm$ 0.50} & 3.16 $\pm$ 1.04 \\
         \bottomrule
    \end{tabular}
    }
    \label{tab:quantitative-sim}
\end{table*}

\subsection{Objective Analysis}
\label{sec:results-quantitative}

\subsubsection{Simulation}
As seen in Table \ref{tab:quantitative-sim}, all residual policies significantly outperform the PC policies across all hold-lengths.
The observed 8\% to 20\% percent improvement indicates that the improved reward function and the driver modelling have a significant impact on the policies.
The mis-specified reward function, $R^{PC}$, leads to policies that perform comparatively with a constant speed policy (OSL). 
The results also indicate that the residual policies are more consistent and stable in their performance when compared to the PC policies due to the consistently low standard deviation in all metrics.
We further analyze the effects of our driver modeling and traits in Appendix \ref{sec:ablation-traits}.

The two policies with trait information, PeRP and TA-RP, perform comparatively with the vanilla residual policy, RP, but only outperform it in the case of $\delta=100$.
This trend indicates that the traits are not extremely informative in providing better advice, when compared to the inferences gained from the real-time experience at every step of the simulation.
This claim is further backed by the PeRP consistently outperforming or performing comparatively with the oracle policy, TA-RP. 
Our findings suggest that the learned latent vectors are better indicators of a drivers' behavior when compared to the ground truth traits.
We posit that this observation can be attributed to the DTI module capturing more nuanced driver behavior that is unrelated to the driver traits. 
Additionally, tests on the driving trajectories conditioned on the driver traits indicate that the driver traits do not affect driving uniformly \ie~there are certain situations where the driver traits are dormant. 
We expand upon the manifestations and effects of the traits in Appendix \ref{sec:appendix-dti}.

\begin{figure}[t!]
    \centering
    \includegraphics[width=\textwidth]{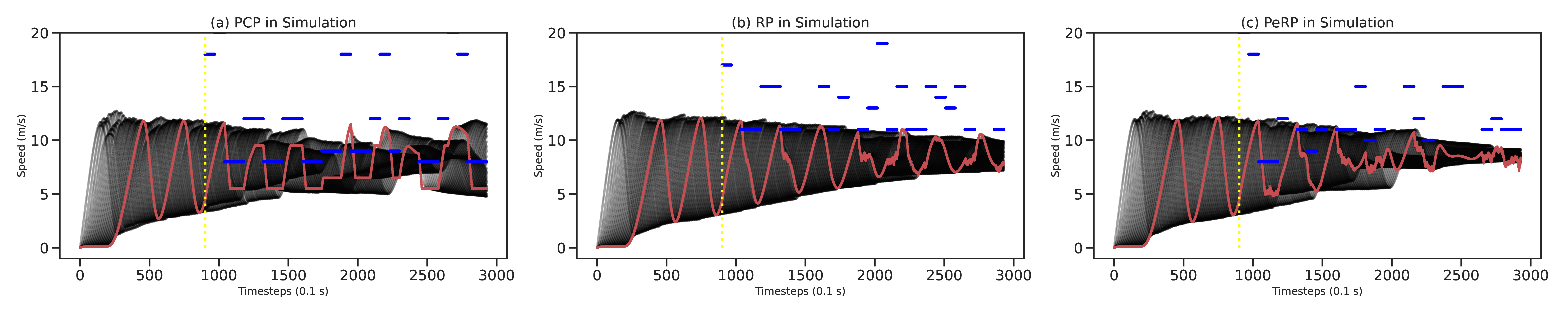}
    \caption{\textbf{Space-Time Graphs in Simulation:} The trajectory of the ego vehicle is shown in \textcolor{red}{red}. The trajectories of the non-ego vehicles are shown in black. Actions advised by the policy are displayed in \textcolor{blue}{blue}. These simulations were conducted for a driver policy who drives 2.5m/s under the advised speed with a hold-length of 70 (7s). The dotted yellow line indicates when the policy begins advising the driver.}
    \Description[Three Speed-time diagrams]{The three diagrams show that the residual policies outperform the baseline policy in simulation.}
    \label{fig:st-sim}
\end{figure}

Figure \ref{fig:st-sim} shows the speed-time diagrams obtained in simulated runs for PCP, RP, and PeRP, respectively\footnote{The individual runs of the policies, such as those shown in Figure \ref{fig:st-sim}, can be random and can have vast differences between them. 
However, the averaged results of 100 evaluation runs shown in Table \ref{tab:quantitative-sim} show that there is a general trend in the policies performance. 
}.
The speed-time diagrams plot the speed of every vehicle over time.
In our diagrams, we indicate the trajectories of the ego vehicle in \textcolor{red}{red} and the trajectories of all other vehicles in black.
The \textcolor{blue}{blue} trajectories indicate the advised speed, and the yellow dotted line (at timestep 900) indicates when the policy begins providing advice to the driver. 
The simulated runs shown in Figure \ref{fig:st-sim} were obtained for a driver with intentional offset intent of -2.5m/s, meaning that the driver policy would intentionally apply an action that is 2.5m/s less than the advised action.
The effect of this intent can be seen clearly in Figure \ref{fig:st-sim} (a) as the ego agent's action is exactly 2.5m/s below the advised action.

As observed in Figure \ref{fig:st-sim} (a), the action advised by PC policies does not adapt to the driver's trait. 
Conversely, we see that both the RP and PeRP provide a higher action value to counter for the driver's intent.
As mentioned in Section \ref{sec:prelim-drawbacks}, we observe (generally) that the $a^{PCP}$ either stays constant at a low speed, or jerks randomly to an incredibility high value, that can cause users to panic.
In contrast, both residual policies provide gradual changes in advice as compared to the PC policy.
However, the residual policies also have occasional slips where the actions are jerky as seen in Figure \ref{fig:st-sim} (b) and (c).
Consequently, we can also observe that the residual policies are able to mitigate congestion more efficiently than the PC policy due to the above factors as evidenced by the convergence of speeds for all vehicles towards the end of the trials.
Furthermore, as seen in Figure \ref{fig:st-sim}, both residual policies gradually increase the recommended speed steadily after a base speed for the user is established, thus mitigating congestion.

\subsubsection{User Study}

\begin{table*}[b!]
    \centering
    \caption{Quantitative results from the User Study}
    \resizebox{\textwidth}{!}{
    \begin{tabular}{l  c c c  c c c  c c c}
        \toprule
         \multirow{2}{*}{\shortstack[l]{Policy \\ Type}} & 
         \multicolumn{3}{c}{$\delta=50~(5\text{s})$} & 
         \multicolumn{3}{c}{$\delta=70~(7\text{s})$} &
         \multicolumn{3}{c}{$\delta=100~(10\text{s})$} \\
         \cmidrule(lr){2-4} \cmidrule(lr){5-7} \cmidrule(lr){8-10} 
         & CF $(\uparrow)$ &$\mu(\uparrow)$& $\sigma(\downarrow)$
         & CF $(\uparrow)$ &$\mu(\uparrow)$& $\sigma(\downarrow)$
         & CF $(\uparrow)$ &$\mu(\uparrow)$& $\sigma(\downarrow)$\\
         \midrule
         PCP &  
         5.99 $\pm$ 0.86 & 6.33 $\pm$ 0.90 & 2.34 $\pm$ 0.86 &
         5.34 $\pm$ 0.61 & 5.55 $\pm$ 0.70 & \textbf{1.74 $\pm$ 0.65} &
         4.82 $\pm$ 0.63 & 5.01 $\pm$ 0.75 & 1.67 $\pm$ 0.64 \\ 
         RP &
         \textbf{7.03 $\pm$ 0.68} & \textbf{7.31 $\pm$ 0.62} & \textbf{2.03 $\pm$ 0.74}  & 
         \textbf{6.56 $\pm$ 0.76} & \textbf{6.88 $\pm$ 0.70} & 2.15 $\pm$ 0.66 &
         6.62 $\pm$ 0.64 & 6.75 $\pm$ 0.55 & \textbf{1.55 $\pm$ 0.87} \\ 
         PeRP & 
         6.23 $\pm$ 0.71 & 6.52 $\pm$ 0.70 & 2.22 $\pm$ 1.21 & 
         6.46 $\pm$ 0.55 & 6.72 $\pm$ 0.61 & 2.00 $\pm$ 0.90 &
         \textbf{6.79 $\pm$ 0.69} & \textbf{7.05 $\pm$ 0.59} & 1.89 $\pm$ 0.59\\
         \bottomrule
    \end{tabular}
    }
    \label{tab:quantitative-user-study}
\end{table*}

Table \ref{tab:quantitative-user-study} shows that the results from the user study with human drivers are in agreement with the results in simulation. 
Specifically, the residual policies consistently outperform the baseline policy as evidenced by the 17\% to 40\% percent improvement in CF. 
Univariate analysis of ANOVA tests on the results obtained in the user study indicated statistically significant differences in the performance of the policies with CF as the dependent variable:
\begin{itemize}
    \item For the hold length, $\delta$, as the independent variable, an F statistic of $4.06$ with $p < 0.05$, indicated a significant effect of each hold-length.
    \item For the policy type (PCP, RP, or PeRP) as the independent variable, an F statistic of $59.01$ with $p < 0.001$, indicated a significant effect of each policy type.
    \item With the combination of $\delta$ $\times$ policy type as the independent variable, an F statistic of $59.11$ with $p < 0.004$, indicated a significant effect of each trained policy across hold lengths.
\end{itemize}
Furthermore, we observed no significant difference in performance based on the order of when the policies were tested during trials across participants.
One-way ANOVA tests with the order of appearance for the policy, and various combinations of the order with other independent variables produced a F statistic of $0.72$ with $p > 0.9$, indicating that \textit{when} participants drove with a particular policy did not affect the outcome of the trial.

\begin{figure}[t!]
    \centering
    \includegraphics[trim=350pt 0cm 0cm 0cm, clip=true, width=\textwidth]{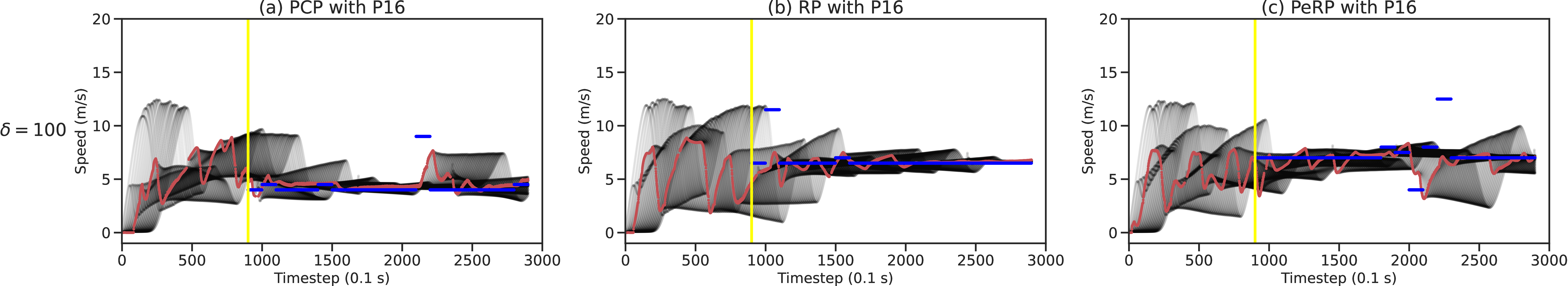}
    \caption{\textbf{Space-Time Graphs for the User Study (P16 with $\delta = 100$):} The trajectory of the participant P16's vehicle is shown in red. Black waves represent the non-ego vehicles. Actions advised by the policy are displayed in blue. Additional figures for other participants are presented in Appendix \ref{sec:more-results}.}
    \Description[Three Speed-time diagrams]{Speed-time diagrams showing that the residual policies perform better than the baseline policies with a user.}
    \label{fig:st-us}
\end{figure}

As the same policies were evaluated in simulation and the user study, with no fine-tuning for the sim-to-real transfer, we can see that our driver modeling  successfully aids in training policies that robust to nuances in deployment settings.
When comparing the speed-time graphs generated in simulation (Figure \ref{fig:st-sim}) and the user study (Figure \ref{fig:st-us}), differences in driving trajectories become clear\footnote{Some observations stated below are drawn from general trends observed in various trajectories from all participants that are not shown here due to space constraints.}.
First, contrasting the warmup periods (areas before the yellow dotted lines) shows that real human trajectories are irregular and not harmonic like the simulated trajectories.
The power that the actions of one vehicle hold over all vehicles in a network is highlighted by the effect of the human driver's irregularities on the simulated cars trajectories.
We observe that humans accelerate linearly and decelerate abruptly at significantly different rates, both with (right side of yellow line) and without (left side of yellow line) advice.
In general, there seem to be significant differences in real driving trajectories and IDM trajectories that merit more investigation into realistic car following policies beyond the current standard.

Secondly, we can see that participants do not follow the advice of the policies exactly.
Both intentional offsets and deviations that arise intrinsically due to the hardware setup can be observed in Figure \ref{fig:st-us}.
In particular, in Figure \ref{fig:st-us} (a), the participant holds a speed higher than that advised speed because they felt that the advised speed was too slow (similar qualitative analysis is presented in Section \ref{sec:results-qualitative} below).
The driving trajectories of participants waver slightly and do not maintain a constant speed. 
We posit that this wavering is due to intrinsic factors that affect the participants ability to control the vehicle such as reaction delays, and experimental hardware factors.
We attribute the robustness of the residual policies toward these new driving behaviors that were not observed during training to our driver modelling.

\subsection{Subjective Analysis}
\label{sec:results-qualitative}
\begin{table*}[t!]
    \centering
    \caption{Qualitative Metrics from the user study}
    \begin{tabular}{l  c c   c c   c c}
        \toprule
         Policy & 
         \multicolumn{2}{c}{$\delta=50~(5\text{s})$} & 
         \multicolumn{2}{c}{$\delta=70~(7\text{s})$} &
         \multicolumn{2}{c}{$\delta=100~(10\text{s})$} \\
         \cmidrule(lr){2-3} \cmidrule(lr){4-5} \cmidrule(lr){6-7} 
         & Raw TLX $(\downarrow)$ & SUS $(\uparrow)$
         & Raw TLX $(\downarrow)$ & SUS $(\uparrow)$ 
         & Raw TLX $(\downarrow)$ & SUS $(\uparrow)$\\
         \midrule
         PCP &  
         \textbf{10.11 $\pm$ 4.43} & \textbf{47.97 $\pm$ 23.88} & 
         \textbf{7.72 $\pm$ 3.87} & \textbf{61.56 $\pm$ 18.88} &
         \textbf{6.00 $\pm$ 1.77} & 65.47 $\pm$ 19.48 \\ 
         RP &
         10.72 $\pm$ 2.95 & \textbf{47.97 $\pm$ 22.06} & 
         9.10 $\pm$ 3.93 & 60.17 $\pm$ 20.41 &
         6.43 $\pm$ 2.92 & \textbf{72.5 $\pm$ 18.95} \\ 
         PeRP & 
         10.33 $\pm$ 3.99 & 45.00 $\pm$ 22.58 & 
         9.64 $\pm$ 3.72 & 56.54 $\pm$ 21.19 &
         7.03 $\pm$ 2.90 & 71.72 $\pm$ 20.37 \\
         \bottomrule
    \end{tabular}
    \label{tab:qualitative-user-study}
\end{table*}

\begin{figure}[t!]
    \centering
    \includegraphics[width=\textwidth]{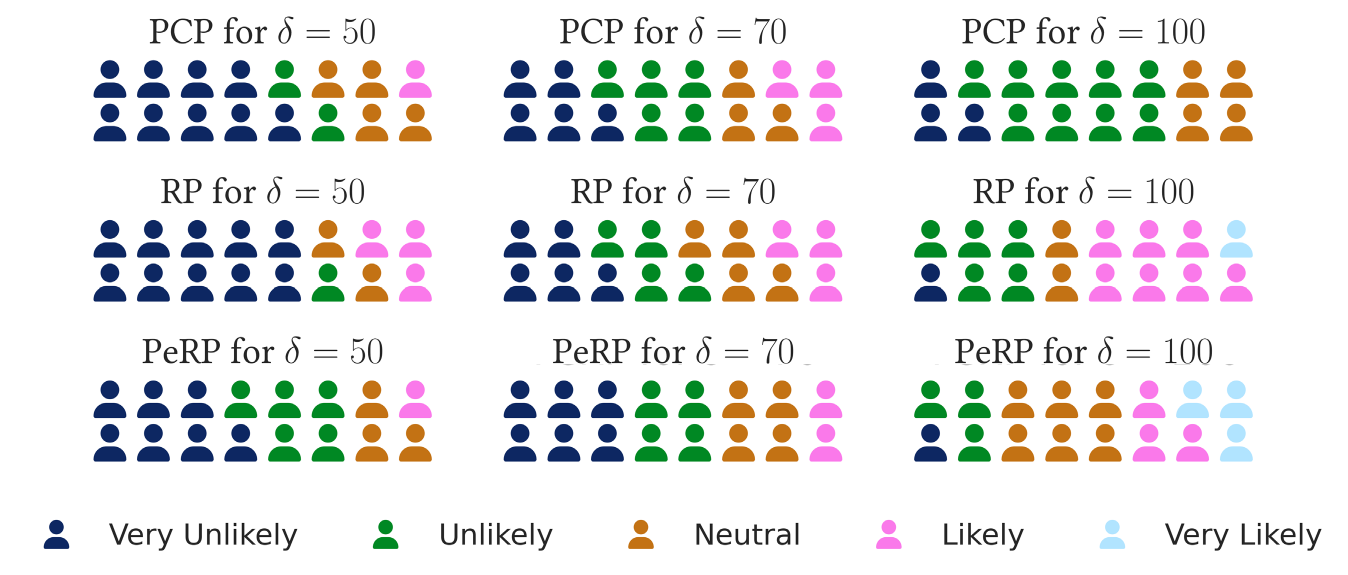}
    \caption{Participant Responses for "How likely are you to use this system?"}
    \Description[Waffle plots]{Waffle plots showing that participants prefer to use PeRP policies and policies with higher hold-lengths.}
    \label{fig:qual-us-histogram}
\end{figure}

The scores obtained from the NASA TLX and SUS questionnaires are presented in Table \ref{tab:qualitative-user-study}. 
Figure \ref{fig:qual-us-histogram} depicts waffle plots showing participant responses to how likely they were to use the different policies in the real world.
The general trend in Table \ref{tab:qualitative-user-study} and Figure \ref{fig:qual-us-histogram} indicates that participants prefer policies with larger hold lengths.
In particular, all policies of a hold length, $\delta_i$, perform significantly better than all policies of smaller hold lengths, $\delta_j < \delta_i$.
Participants particularly remarked that for policies with shorter hold-lengths, the time for them to react to the advice was too short before the policy would change its advice. 
This phenomenon made participants frustrated with the policies and the advice that was given to them.

While the trends in the subjective metrics show a preference for PC policies, participants voiced concerns that the PC policies were particularly frustrating to use.
The SUS scores do not immediately reflect this observation as the frustration is only one aspect of the questionnaire. 
Additionally, the difference in SUS scores for the policies is not statistically significant to counter the participants qualitative feedback.
A major cause of users' frustration with PC policies lay in its behavior to either suggest slow constant actions for the whole trial period, or to change the advice between extremes at a high frequency.
The slow constant actions made the drivers feel as though they were the cause of congestion on the road.
\begin{displayquote}
\textit{It was going too slow. I feel like I was the one who was making this congestion, so I was a little stressed out when there was[sic] no cars in front of [me]. ... It was just frustrating.} - P02 for PCP$_{\delta=100}$
\end{displayquote}

Participants touted the conservative nature of the PC policies when they provided a slow constant action as safe.
Multiple participants remarked that this behavior made the policy ``easy to follow" (P10 for PCP$_{\delta=100}$), which is reflected in the results shown in Table \ref{tab:qualitative-user-study} as all PC policies scored the lowest TLX score.
However, they also commented that the policy was ``a little too safe" (P14 for PCP$_{\delta=50}$) and that ``the recommendation doesn't make sense." (P12 for PCP$_{\delta=50}$) for PC policies as it did not follow the flow of traffic.

In contrast, the residual policies were ``more reactive" (P01 for PeRP$_{\delta=70}$) and seemed to be ``more reasonable" (P05 for RP$_{\delta=100}$).
The residual policies also seemed to understand the flow of traffic in the environment better and provide guidance accordingly.
\begin{displayquote}
\textit{I think this was ... intelligently guiding me. Sometimes I'm a little bit ... conservative, afraid of collision, but ... it actually made me more, go faster... I was stopping. But then it was guiding me to go a little bit faster. I think that might mitigate the whole congestion problems... I think it could be kind of more intelligent than human driving and can help the congestion mitigation} - P02 for PeRP$_{\delta=100}$
\end{displayquote}

Generally, participants disliked when the advice provided by the policies jumped, \ie~differed by a significant amount in consecutive hold periods, for all policies and hold lengths. 
Jumps greater than 10mph ($\sim$ 4m/s) were particularly disliked, while jumps less than that were tolerable as they were ``easy to adjust to" (P14 for PeRP$_{\delta=100}$).
We observed that the residual policies had smaller jumps ($<$ 20mph $\sim$ 10m/s), when compared to the PC policies ($>$ 50mph $ \sim$ 20m/s).
Reacting to the large jumps in PC policies placed more load on the participants and made them uncomfortable and left them unsatisfied:
\begin{displayquote}
\textit{I think I was slamming down on the accelerator a lot of times just get up to speed... I think it was just very, very jumpy with little speed targets. So, I was never very satisfied with the speed I was going out to.} - P05 for PCP$_{\delta=50}$.
\end{displayquote}
In general, participants felt that the presence of jumps in the advised actions pushed them towards wanting to ignore the advice from the policy as a whole.
Additionally, participants also indicated a dislike for being told a particular speed to follow.
They preferred being offered suggestions on how to change their speed (\eg~speed up, maintain speed, slow down) as they would find those easier to follow.


Interestingly, we note that there is a disparity between the quantitative results and the qualitative results when comparing the inferences from Tables \ref{tab:quantitative-user-study} and \ref{tab:qualitative-user-study}.
While the residual policies are objectively better at mitigating congestion they perform comparatively with the baseline policy in the subjective metrics.
The subjective participant sentiments indicated a clear preference for the residual policies as they were less frustrating to use and matched what participants would do in real life.
We hope to perform future work to standardize the metrics used to compare human-compatible congestion mitigation policies to be more cohesive.
Detailed inferences obtained from the subjective semi-structured interviews with regards to how the behaviors of the policies affected drivers are discussed in supplementary material\footnote{Uploaded as additional material} as they are out of scope for this article. 
We encourage interested readers to refer to the attached document as we discuss driver sentiments towards various behaviors of the policies (\eg~ changes in advice, trust in the system) and the user interface, and provide recommendations for design improvements in real-time advisory policies.

Considering the above quantitative and qualitative observations, we conclude that our residual policies successfully mitigate or dampen traffic congestion while also being more human-compatible.
However, our methods are not without their limitations.
We discuss the limitations and possible avenues for future work in the next section.

\section{Limitations and Future Work}
\label{sec:limitations}
The noteworthy limitations of our work are discussed below.

First, as the policies are trained with RL, training stability is heavily dependent on the policies' balance of exploration and exploitation. 
This dependency necessitates the need to train multiple policies which can be cumbersome and hinder reproducibility. 
We are encouraged by recent advances in improving reproducibility of RL methods and hope to apply the techniques here~\cite{khetarpal2018reevaluate}.
Second, as our policies are residual policies, they are heavily limited by the base policy. 
Therefore, some unwanted artifacts of the base policy propagate into the residual policy which are hard to unlearn, \eg~jumps between consecutive actions (See Appendix \ref{sec:more-results}).
As our method is agnostic to the base policy used, the residual policies could also be constructed for use with classical control policies for congestion mitigation.
Third, TfS policies that have explicit emphasis on providing smoother action changes by design, rather than through reward shaping could also be trained to produce more human-compatible policies.

Additionally, our policies are only tested on the simple ring road without lane changes. 
Further testing on other networks such as the open ring or intersection environments~\cite{flow}, and the inclusion of lane changing would provide further credence to our method.
We are encouraged that our policies would perform well in these settings as PC policies have been provably robust to these changes in environment~\cite{pcp, li2023stabilization}.
In principle, increasing the penetration of our policies from one vehicle to multiple vehicles also holds promise. 
Testing and evaluating the penetration power of such a design could be fruitful in addressing congestion mitigation.
Lastly, as our user study was conducted in a driving simulator environment, the sim-to-real gap between the deployment of the policies in simulation and on a real-world road necessitates naturalistic studies.
We are encouraged by studies performed by \citet{stern2018dissipation, nice2021can, nice2023enabling, hayat2023traffic} that our methods can be tested in an easier fashion as we would not require external hardware beyond the sensors on the ego vehicle. 

\section{Conclusion}
\label{sec:conclusion}
In this article, we introduce residual policies that can aid in mitigating traffic congestion while being human-compatible.
Our novel class of residual policies are based on Piecewise Constant Policies and address the drawbacks associated with PC policies through designing an improved reward function and robust driver modeling for instruction following.
Our policies are trained in simulation using SUMO with our novel driver modeling framework.
We showcase the performance of our policies through experiments in simulation and through a driving simulator user study (N=16) performed using CARLA.
The results in simulation and through our user study highlight the use of cooperative advisory systems, and driver sentiments towards these systems to uncover avenues for improvement.
Our work showcases the potential of residual RL policies and trait inference methods in not only reducing congestion on the roads, but also in positively impacting driver experiences, and the environment through reducing emissions.

\begin{acks}
The authors would like to thank all the participants in the user study for their participation.
We would also like to thank Kristina Miller, Sirui Li, and Dr. Jeongyun Kim for their support and assistance with various aspects of this work.
\end{acks}

\bibliographystyle{ACM-Reference-Format}
\bibliography{bibliography}

%
\appendix
\section{Training}
\label{sec:appendix-training}
We employ a warm-up period of $W$ timesteps in each rollout to train and evaluate (in simulation only) all policies. 
As seen in Algorithm \ref{alg:rp}, during the warm-up period, all vehicles are controlled by IDM to generate stop-and-go traffic waves. 
Following the warm-up period, the control of the ego vehicle is determined by the driver policy acting on the advised action. 
We set $W=900$ and train for a horizon of $H=3000$ for all policies.

We train and evaluate policies with a learning rate of $1e-4$ and $\gamma=0.99$.
During training and evaluation, all actions advised to the driver are clipped between $0$ and $A_{max}$.
All policies were parameterized by a two-layer multi-layered perceptron with $64$ nodes each and trained with TRPO~\cite{schulman2015trpo}.

\textbf{PC Policies:}
We train for $1000$ steps for each hold length and with $A_{max} = 35$. 
In order to compare our policies with those with a constrained action space, we also examine PC policies and residuals that are trained with $A_{max} = 10$ and discuss their performance in Appendix \ref{sec:ablations}.
All PC policies were trained on the MIT SuperCloud with an Intel Xeon Platinum 8260 processor and 4 CPUs~\cite{reuther2018interactive}.

\textbf{Residual Policies:}
All residual policies, RP, PeRP, and TA-RP were trained using the same base PC policy for each hold-length.
Training was performed with $n_{actions} = 10, \epsilon=1$ for $1000$ steps.
We emphasize that training all residual policies is $2\times$ faster than training the PC policies, and TfS residual policies.
All training was performed on a machine with an Intel i7-10700 processor and a single Nvidia 3080 GPU.
Lastly, we set the hyperparameters of the reward function $\alpha_{speed} = 1, \alpha_{headway} = 1$, and $\alpha_{action} = 0.5$

\textbf{DTI Models:}
We train two DTI models to be used by our PeRPs to capture the response time and intentional action deviation traits, respectively.
A custom dataset was curated in simulation where the driver policy followed advice given by PC policies for each delta.
The dataset comprised of $100000$ driving trajectories with equal distributions among all the driver traits considered.
We performed a $80-20$ train-test split on the custom dataset.

Both models were trained for 100 epochs with a learning rate of $1e-3$ with a latent size of $2$, a batch size of $16$, $\beta_{\text{recon}} = 1$ and $\beta_{\text{KL}}=1e-6$.
We use a single layer LSTM with hidden and cell sizes of $32$ as our base recurrent network for both the encoder and decoder as shown in Figure \ref{fig:method-overview}.

\section{Driver Trait Inference}
\label{sec:appendix-dti}

\begin{figure}[t!]
    \centering
    \begin{minipage}{0.49\textwidth}
        \centering
        \includegraphics[width=\textwidth]{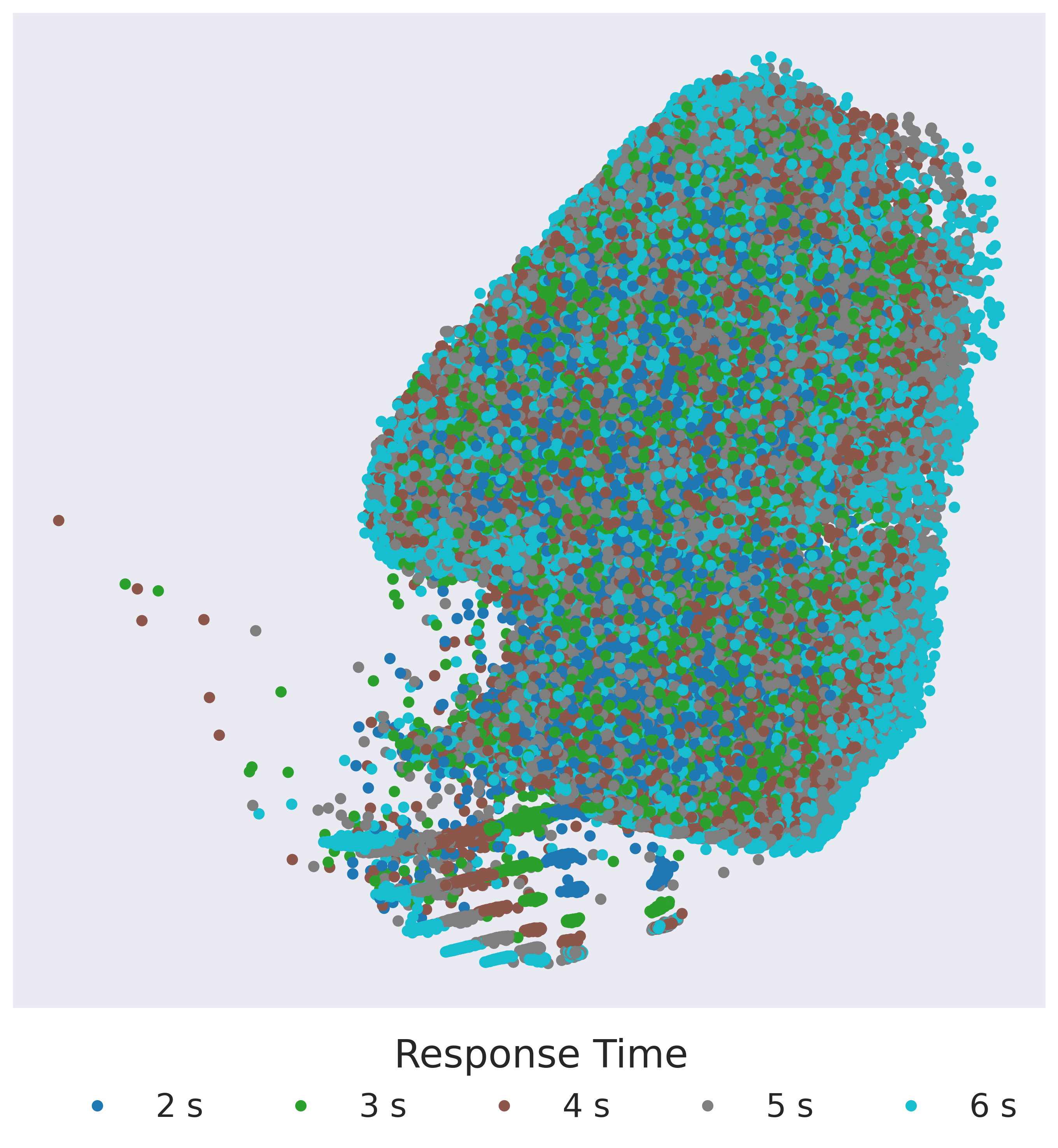}
        \caption{Latent space for Response Time}
        \Description{A figure showing different colored points for the response time. Some points from clusters separated by the time, others are all combined into a big cluster.}
        \label{fig:latent-response-time}
    \end{minipage}\hfill
    \begin{minipage}{0.49\textwidth}
        \centering
        \includegraphics[width=\textwidth]{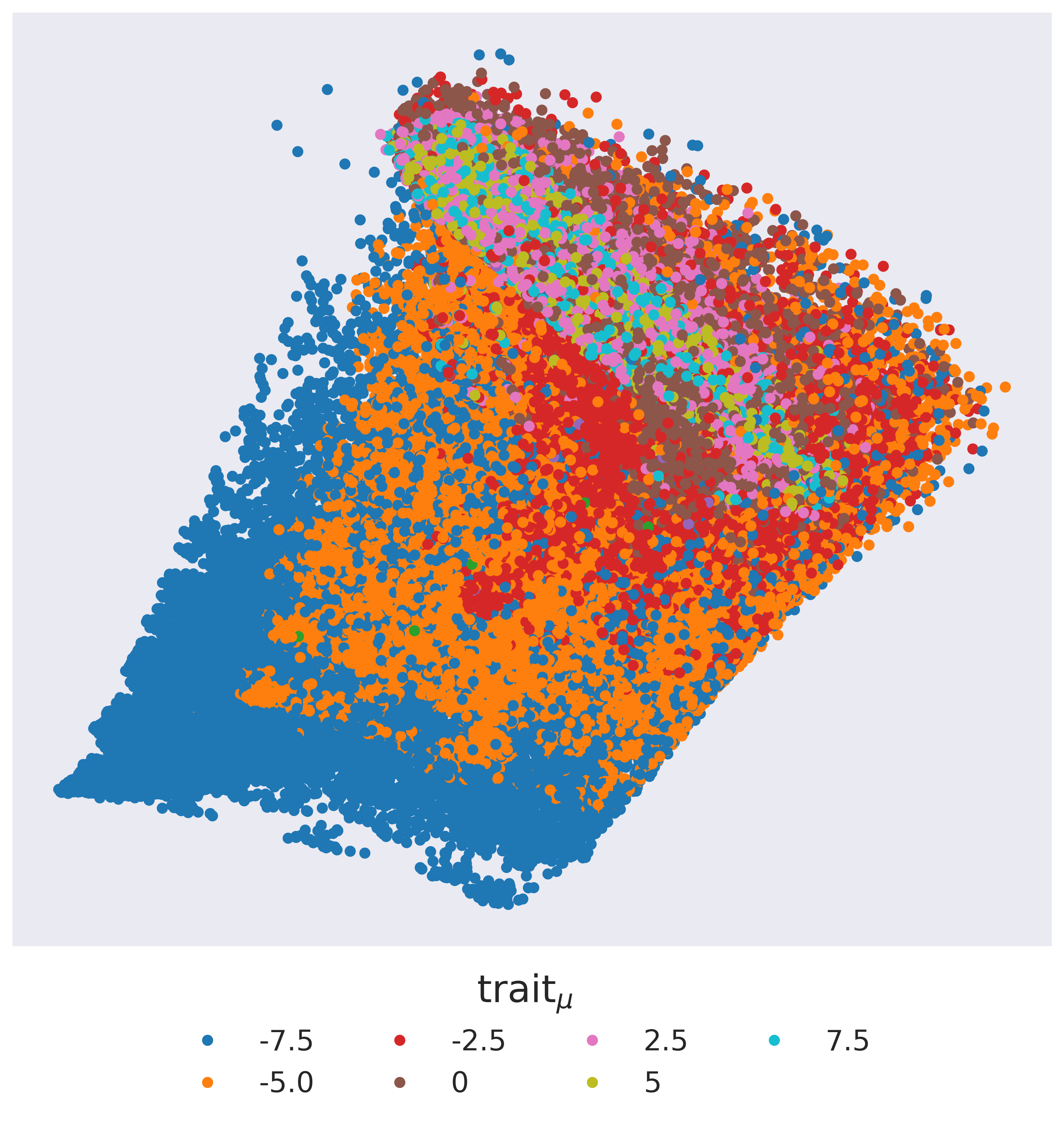}
        \caption{Latent space for Intentional Instruction Offset}
        \Description{A figure showing different colored points for the different offsets, separated by the offset. Some points are clustered by their colors, others are combined into a one big cluster.}
        \label{fig:latent-intentional-deviation}
    \end{minipage}
\end{figure}
Figures \ref{fig:latent-response-time} and \ref{fig:latent-intentional-deviation} depict the latent spaces for the two DTI models that were trained and used for the PeRPs.
Recall that our DTI models take as input a simulated driving trajectory and output a latent vector representing the driver's inferred trait.
In both figures shown here, each point represents a simulated driver trajectory, and the color of the point represents the corresponding trait.

Both latent spaces show clear separations in the driver traits, indicating that the different traits and the driving trajectories produced by drivers with those traits can be captured.
However, there also exist significant overlaps in the latent space \eg~ the large blob of mixed points in Figure \ref{fig:latent-response-time}.

As mentioned in Section \ref{sec:results-quantitative}, the driver traits can sometimes be dormant.
For example, the driver traits would not manifest themselves in the driving trajectory in the following two conditions:
\begin{enumerate}
\item If a driver takes 5 seconds to respond to instructions, but the advised instructions does not change for multiple hold length periods.
\item If the driver tends to intentionally drive 5m/s over the provided advice but the advised speed is such that driving any faster would lead to collisions, the driver physically cannot exercise their intentional offset trait without causing a collision.
\end{enumerate}
Such scenarios are ubiquitous in everyday driving and are hence expected to occur.
We are encouraged that these scenarios also occur in our simulated driving and are readily visible in our learned latent spaces. 
A direct consequence of the dormant nature of the traits is seen in Tables \ref{tab:quantitative-sim-driver-intent} and \ref{tab:quantitative-sim} in the under-performance of the trait aware residual policy, TA-RP, when compared to the other residual policies.

Note that while the trajectories that are clumped together in right side in Figure \ref{fig:latent-response-time} arise from different traits, their grouping still indicates similarities in the driving trajectories.
We argue that the these learned groupings, that cannot be attributed to an immediate driver trait, are more informative about the driver's behavior in following driving instructions.
The performance of the DTI enhanced policy, PeRP, and the vanilla residual policy, RP, when compared to the performance of TA-RP add credence to this claim.

\subsection{Driver Trait Inference on Human driver trajectories}
\label{sec:dti-st}
\begin{figure}[t!]
    \centering
    \includegraphics[width=\textwidth]{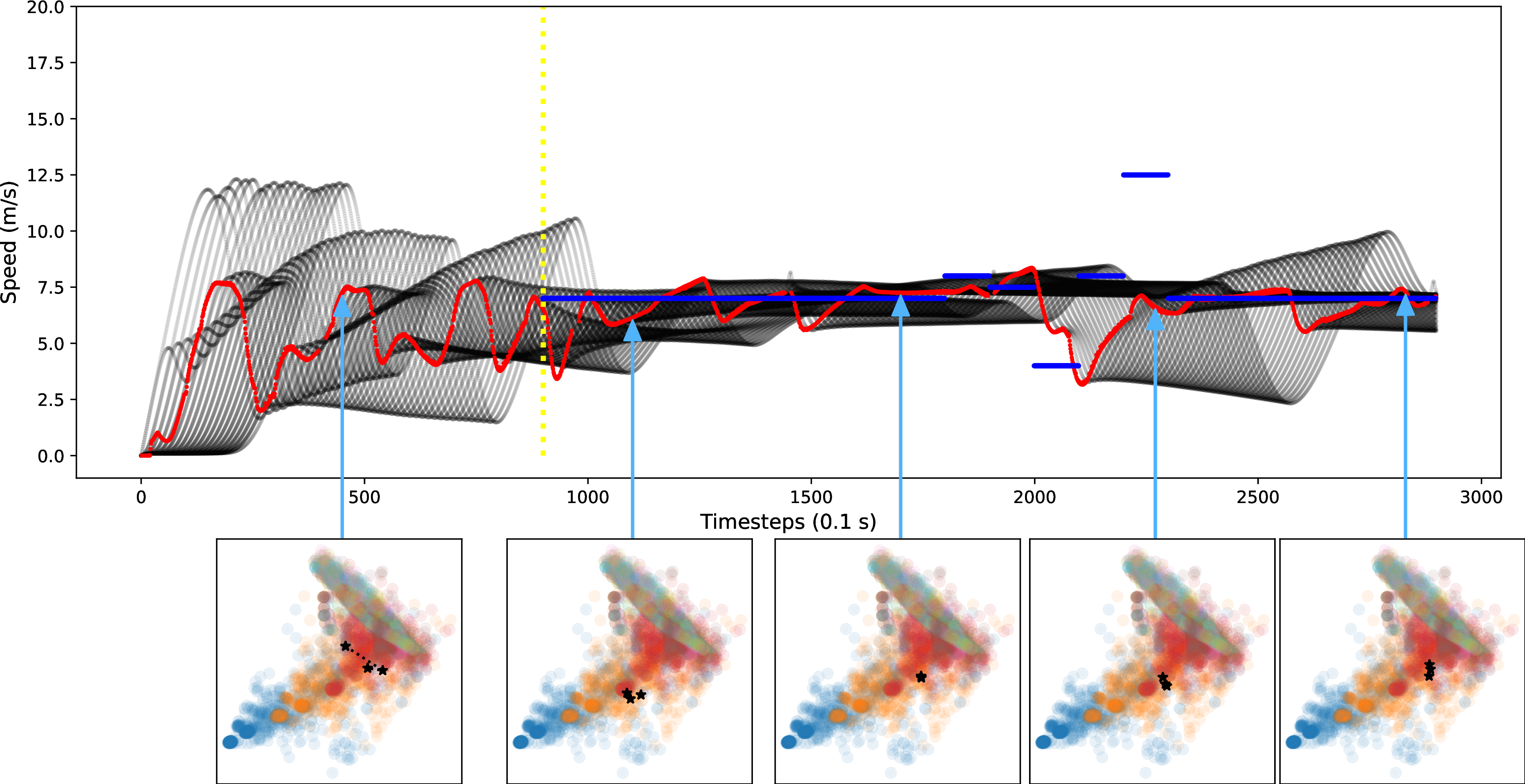}
    \caption{Plotting the predicted latent trait vector at different points of a trial for P16 for PeRP with $\delta=100$. The insets plot the evolution of the predicted latent vectors for the intentional offset trait for the last three hold-length periods. The colors in the inlay plot represent the different driver traits as in Figure \ref{fig:latent-intentional-deviation}, with the predicted latents in black. }
    \Description{Insets of the latent vectors on a speed-time diagram. Five insets show how the different points move.}
    \label{fig:driver_traits_st}
\end{figure}

Figure \ref{fig:driver_traits_st} shows the predicted latent for the driver traits at different points during a user trial. 
Note that we only plot a subset of the latent points from Figure \ref{fig:latent-intentional-deviation} in the insets. 
We plot the last three predicted latent points for the trajectory shown at the time corresponding to the speed and advice as indicated by the \textcolor[HTML]{00B6FF}{blue} arrows.
We observe that the predicted latent vectors change throughout the trial period. 
The intentional offsets are predicted to be large and negative at the beginning of the trial.
Once the driver is closer to the optimal advised speed for the network, the latent vectors stabilize at a smaller offset. 
As the driver speeds deviate from the advised speed, the latent vectors readjust appropriately.
As noted earlier, the DTI models learn differing traits than those expected due to the dormancy expressed by some of the traits.

\section{Additional Results Figures}
\label{sec:more-results}
\begin{figure}[]
    \centering
    \begin{subfigure}[t]{\textwidth}
    \includegraphics[trim=320pt 0cm 0cm 0cm, clip=true,width=\textwidth]{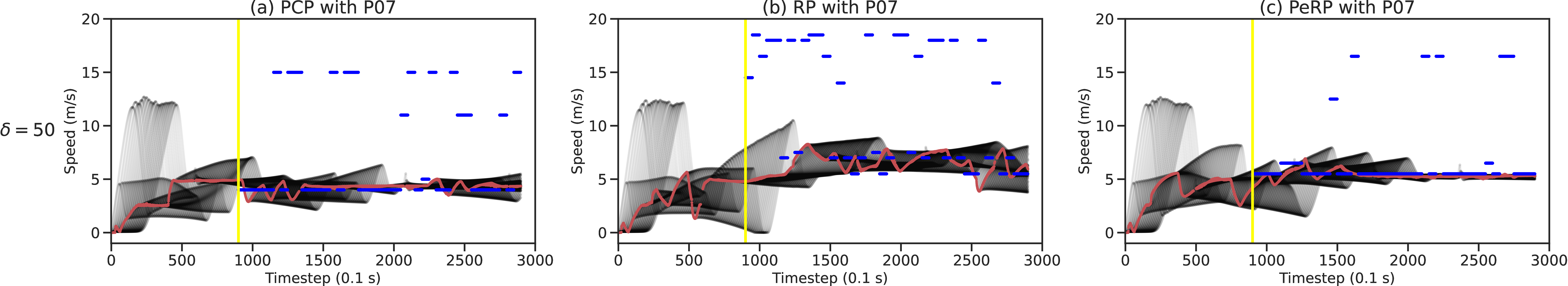}
    \caption{Speed-time diagram for P07 with $\delta=50$}
    \end{subfigure}
    \begin{subfigure}[t]{\textwidth}
    \includegraphics[trim=320pt 0cm 0cm 0cm, clip=true,width=\textwidth]{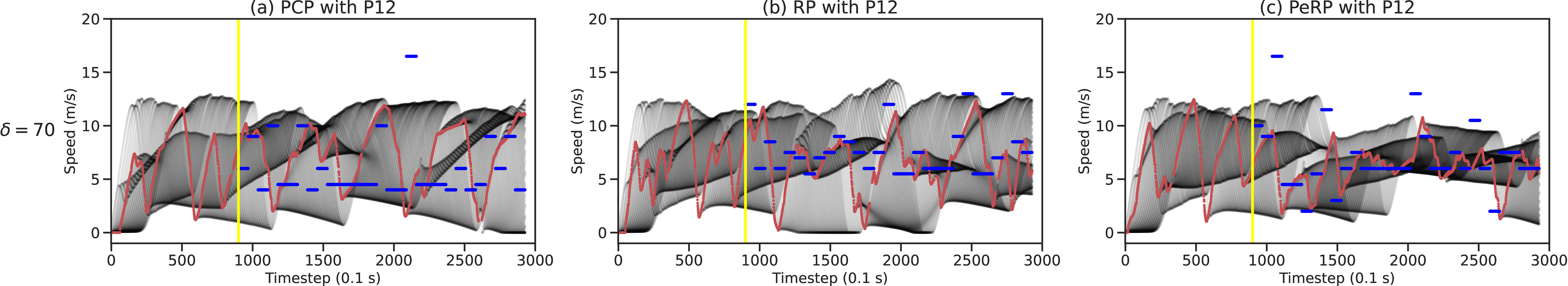}
    \caption{Speed-time diagram for P12 with $\delta=70$}
    \end{subfigure}
    \begin{subfigure}[t]{\textwidth}
    \includegraphics[trim=350pt 0cm 0cm 0cm, clip=true,width=\textwidth]{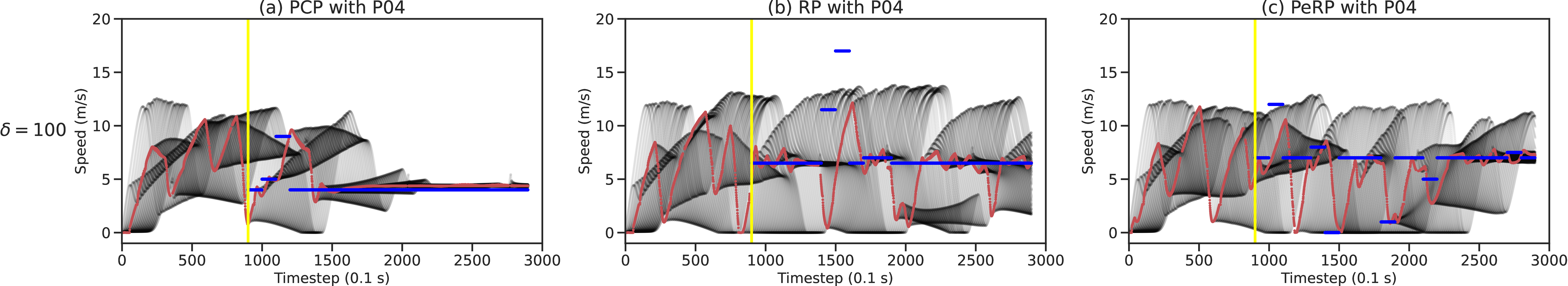}
    \caption{Speed-time diagram for P04 with $\delta=100$}
    \end{subfigure}
    \caption{Additional Space-Time Graphs from the User Study}
    \Description{Speed-time diagrams showing the performance of the different policies in the user study for three users are different hold-lengths.}    
    \label{fig:st-us-more}
\end{figure}

Figure \ref{fig:st-us-more} shows more examples of speed-time graphs from the user study.
These speed-time graphs show more characteristics that we observe with the policies. 
Particularly, Figure \ref{fig:st-us-more} (i)(a) shows that the baseline PC policies oscillate between speeds that are too slow and too fast. In this particular case, P07 opted to ignore the faster recommendations.
We also observe that some of these large jumps appear in the residual policies - indicating that the residual cannot completely unlearn non-ideal actions in the base policy.
The figures also show how the residual policies try to offset the PC policy action: 
(1) The effect of $R^{RP}$ in reducing the difference between continuous recommended actions is shown in the smaller jumps (\eg~Figure \ref{fig:st-us-more} (i)(b) vs. Figures \ref{fig:st-us-more} (i)(a)); 
(2) The recommended actions for subplots (b) and (c) are faster than those in Figures subplots (a) at a baseline level, leading to faster speeds overall while still mitigating congestion.
The figures also show how participants can have a hard time maintaining a constant action (\eg~Figure \ref{fig:st-us-more} (i)), and the variability in how different drivers react to the advice when comparing their speed profiles.

\renewcommand\thesubfigure{\roman{subfigure}}

\begin{figure}[]
    \centering
    \begin{subfigure}[t]{\textwidth}
    \centering
    \includegraphics[trim=0cm 0cm 0cm 0cm, clip=true, height=0.2\textheight]
    {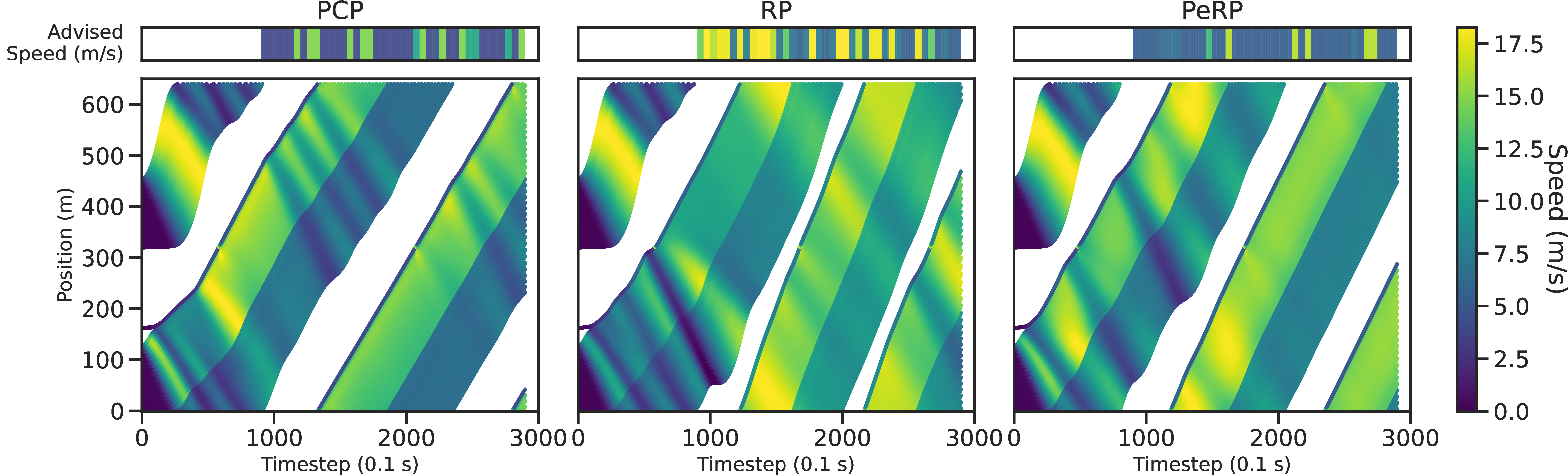}
    \caption{P07 with $\delta=50$}
    \end{subfigure}
    \begin{subfigure}[t]{\textwidth}
    \centering
    \includegraphics[trim=0cm 0cm 0cm 0cm, clip=true, height=0.2\textheight]
    {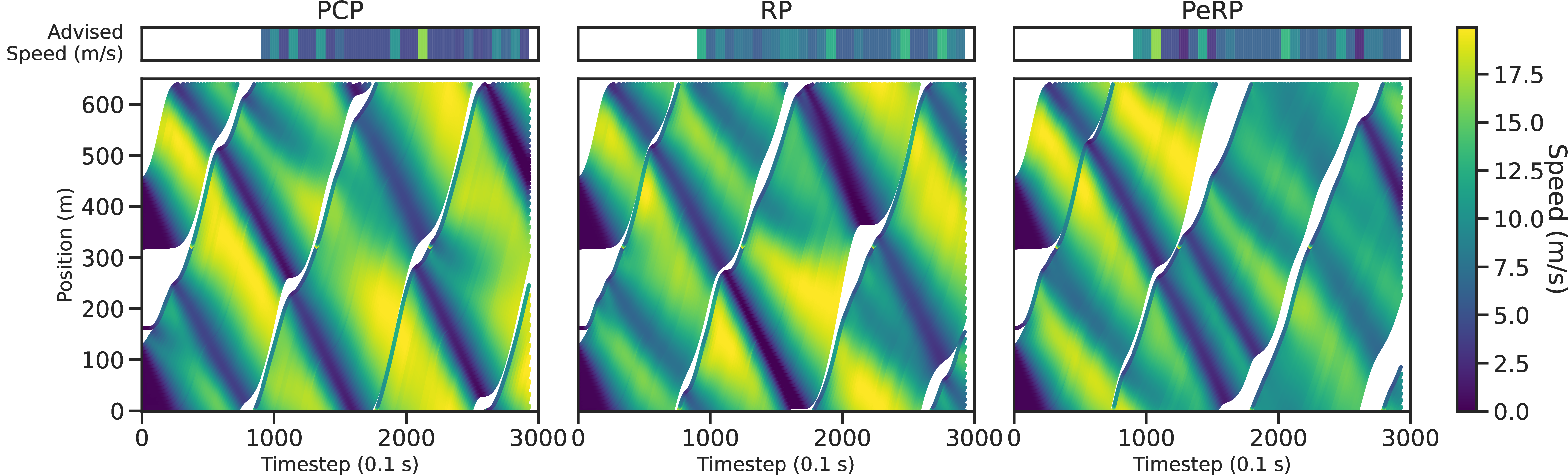}
    \caption{P12 with $\delta=70$}
    \end{subfigure}
    \begin{subfigure}[t]{\textwidth}
    \centering
    \includegraphics[trim=0cm 0cm 0cm 0cm, clip=true, height=0.2\textheight]
    {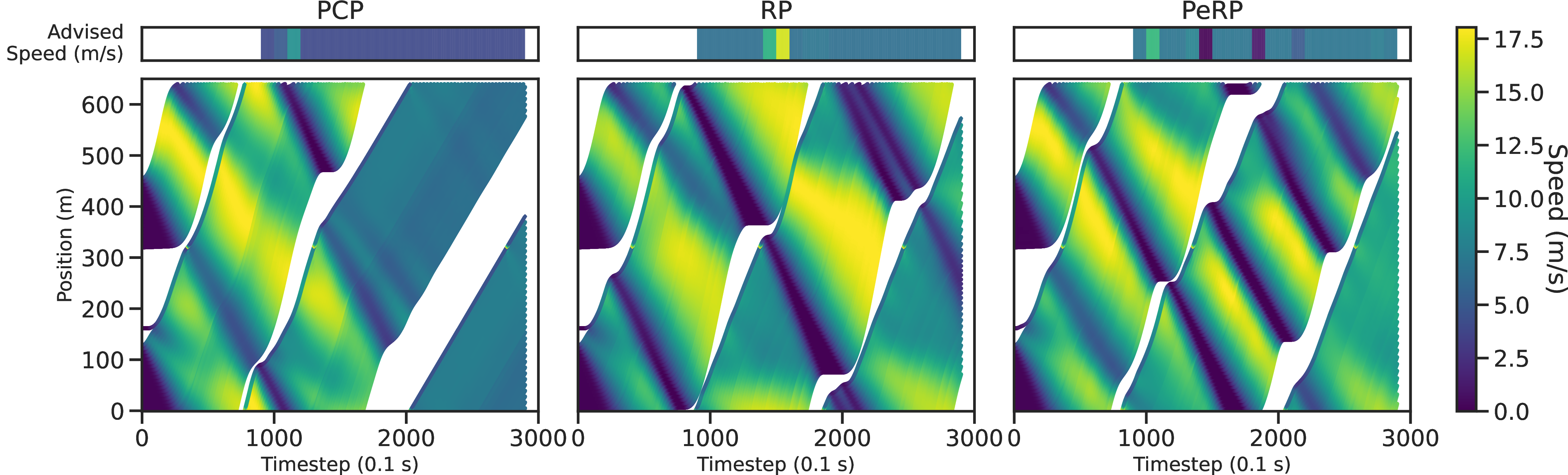}
    \caption{P04 with $\delta=100$}
    \end{subfigure}
    \begin{subfigure}[t]{\textwidth}
    \centering
    \includegraphics[trim=0cm 0cm 0cm 0cm, clip=true, height=0.2\textheight]
    {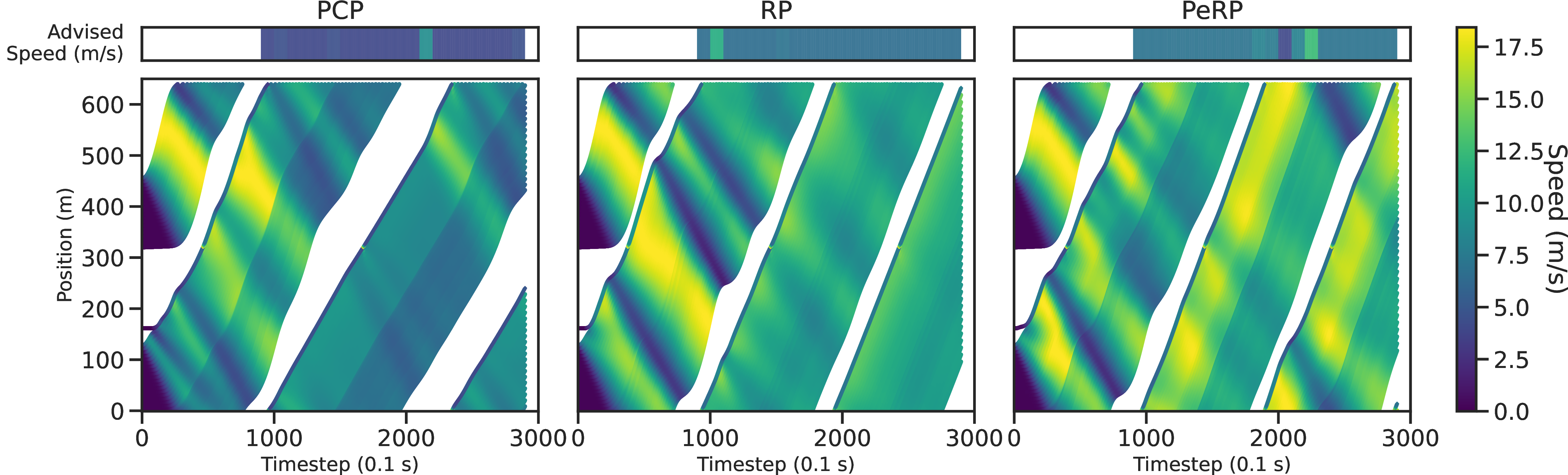}
    \caption{P16 with $\delta=100$}
    \end{subfigure}
    \caption{\textbf{Position-Time Graphs from the User Study}. Each plot shows the position of the vehicles over time, with the color indicating the speed of the vehicle. The darkest line (often the left most) indicates the attributed of the ego vehicle. The bar plots above show the advised speed by the policies.}
    \Description[Four position-time diagrams]{Four position-time diagrams showing the performance of the different policies in the user study.}
    \label{fig:pt-us}
\end{figure}

Alternately to the speed-time diagrams, we can observe similar observations in the position-time diagrams (Figure \ref{fig:pt-us}).
These graphs are an alternate to the speed-time diagrams.
The effect of policies that mitigate congestion is more clearly visible by the absence of the gradient waves in the plots. 
We see that our residual policies lead to brighter (faster), smooth plots with lower white space between the trajectories indicating a smoother flow.
Comparatively the PC plots show uniforms speeds but have more white space (indicating the presence of congestion induced by the ego driver).
Furthermore, the plots also show that all residual policies lead to faster speeds and have more uniform changes in the advised speed (smoother gradients in the top advised speed bars).

\section{Ablation Studies}
\label{sec:ablations}
\begin{table*}[t!]
    \centering
    \caption{Results comparing the performance of our policies with the baselines in simulation for 100 interations with a smaller $A_{max}$}
    \resizebox{\textwidth}{!}{
    \begin{tabular}{l  c c c  c  c c c  c  c c c}
        \toprule
         \multirow{2}{*}{\shortstack[l]{Policy \\ Type}} & 
         \multicolumn{3}{c}{$\delta=50~(5\text{s})$} && 
         \multicolumn{3}{c}{$\delta=70~(7\text{s})$} &&
         \multicolumn{3}{c}{$\delta=100~(10\text{s})$} \\
         \cmidrule{2-4} \cmidrule{6-8} \cmidrule{10-12} 
         & CF $(\uparrow)$ &$\mu(\uparrow)$& $\sigma(\downarrow)$&
         & CF $(\uparrow)$ &$\mu(\uparrow)$& $\sigma(\downarrow)$&
         & CF $(\uparrow)$ &$\mu(\uparrow)$& $\sigma(\downarrow)$\\
         \midrule
         PCP &  
         6.04  $\pm$ 2.34 & 6.10  $\pm$ 2.58 & \textbf{1.61  $\pm$ 1.17} &&
         5.66  $\pm$ 2.46 & 5.71  $\pm$ 2.72 & \textbf{1.54  $\pm$ 1.10} &&
         6.11  $\pm$ 2.14 & 6.16  $\pm$ 2.43 & \textbf{1.65  $\pm$ 1.21} \\
         TA-RP   & 
         7.02  $\pm$ 0.86 & 7.38  $\pm$ 0.89 & 2.36  $\pm$ 0.49 &&
         6.88  $\pm$ 1.18 & 7.18  $\pm$ 1.31 & 2.26  $\pm$ 0.98 &&
         7.01  $\pm$ 1.07 & 7.17  $\pm$ 1.29 & 1.77  $\pm$ 0.95 \\
         RP &
         7.24  $\pm$ 0.60 & 7.57  $\pm$ 0.76 & 2.38  $\pm$ 0.92 && 
         \textbf{7.32  $\pm$ 0.48} & \textbf{7.71 $\pm$ 0.55} & 2.61  $\pm$ 0.81 && 
         7.49  $\pm$ 0.52 & 7.77  $\pm$ 0.65 & 2.23  $\pm$ 1.05 \\
         PeRP & 
         \textbf{7.48  $\pm$ 0.40} & \textbf{7.79 $\pm$ 0.55} & 2.39  $\pm$ 1.02 && 
         7.28  $\pm$ 0.59 & 7.63  $\pm$ 0.68 & 2.44 $\pm$ 0.88 && 
         \textbf{7.52  $\pm$ 0.25} & \textbf{7.87 $\pm$ 0.35} & 2.53  $\pm$ 0.95 \\
         \bottomrule
    \end{tabular}
    }
    \label{tab:quantitative-sim-3}
\end{table*}

\subsection{Effect of a smaller $A_{max}$}
\label{sec:ablation-lower-amax}
The performance of the policies with a more limited action space ($A_{max} = 10$) is shown in Table \ref{tab:quantitative-sim-3}.
All observations discussed in Section \ref{sec:results-quantitative} can be noticed in policies in this case as well.
While the policies presented here are capable of achieving lower standard deviations in speed, $\sigma$, we observed that they break when paired with a human driver during out pilot study.
The under-performance and innate in-flexibility for suggesting higher speeds contributed to our decision to use the less constrained policies in our user study.

\subsection{Analysis of policies conditioned on the traits}
\label{sec:ablation-traits}
Table \ref{tab:quantitative-sim-driver-intent} shows the performance of our policies and the baselines conditioned on the intentional instruction offset trait.
Metrics for the best performing policies are presented in bold in the table. 
No significant deviations in metrics were observed for conditional observations on the response time trait.

In general, we still observe that the residual policies outperform the baseline policy across all traits. 
However, micro-trends can be observed based on the mean intentional offset from the advised action.
The residual polices significantly outperform the baseline policy for all negative offsets from the advised actions \ie~the residual policies are incredibly robust to drivers driving slower than the advised action.
This observation is confirmed by the superiority of \textit{all} residual policies over the baseline policy across all conditions.
As the offsets from the advised speed turn positive \ie~ the driver is driving at a speed higher than the advised speed, the residual policies only slightly outperform the baseline.
It is however likely that this observation is due to the dormancy of the positive intentional offset traits compared to the negative intentional offset traits as discussed in Section \ref{sec:appendix-dti}.

The effect of road network parameters, such as the number of cars and the road length, that prevent drivers from safely driving at higher speeds can be observed in the results for the speeding traits (trait$_\mu=\{2.5, 5, 7.5\}$).
In particular, we notice that the standard deviations in speed increases as the offsets from the advised action increase for all evaluation scenarios.
We conjecture that modifying the road parameters could bring out stronger observations and differences between the policies when drivers over-speed.

Table \ref{tab:quantitative-sim-driver-intent} also shows that PC policies are unable to adapt to the drivers' traits as evidenced by the significant spread in speeds across the trait means and larger variations in the standard deviations for all metrics.
In contrast, the residual policies perform more consistently.
The residual policies still under-perform for the most extreme under-speeding case (trait$_\mu = -7.5$) but significantly improve over the base policy.
Their under-performance can be wholly attributed to the use of the PC policies as the base policy.

\begin{table*}[t!]
    \centering
    \caption{Performance of the policies in Simulation conditioned on the driver intent trait}
    \resizebox{\textwidth}{!}{
    \begin{tabular}{c l  c c c  c  c c c  c  c c c}
        \toprule
         \multirow{2}{*}{\shortstack[l]{Driver \\ Intent}} & \multirow{2}{*}{\shortstack[l]{Policy \\ Type}} & 
         \multicolumn{3}{c}{$\delta=50~(5\text{s})$} && 
         \multicolumn{3}{c}{$\delta=70~(7\text{s})$} &&
         \multicolumn{3}{c}{$\delta=100~(10\text{s})$} \\
         \cmidrule{3-5} \cmidrule{7-9} \cmidrule{11-13} 
         & & CF $(\uparrow)$ &$\mu(\uparrow)$& $\sigma(\downarrow)$&
         & CF $(\uparrow)$ &$\mu(\uparrow)$& $\sigma(\downarrow)$&
         & CF $(\uparrow)$ &$\mu(\uparrow)$& $\sigma(\downarrow)$\\
         \midrule
        \multirow{4}{*}{-7.5}& PCP & 
         3.34 $\pm$ 0.86 & 3.89 $\pm$ 0.90 & 3.59 $\pm$ 0.49 &&
         1.73 $\pm$ 0.25 & 1.98 $\pm$ 0.35 & \textbf{1.83 $\pm$ 0.46} &&
         1.04 $\pm$ 0.50 & 1.29 $\pm$ 0.72 & 2.08 $\pm$ 1.03 \\ 
         & TA-RP &
         5.68 $\pm$ 0.52 & 6.11 $\pm$ 0.56 & 2.67 $\pm$ 0.33 && 
         3.78 $\pm$ 0.55 & 4.24 $\pm$ 0.57 & 2.90 $\pm$ 0.30 && 
         5.75 $\pm$ 0.10 & 5.83 $\pm$ 0.28 & 1.32 $\pm$ 0.51 \\ 
         & RP &
         \textbf{6.75 $\pm$ 0.23} & \textbf{7.17 $\pm$ 0.24} & \textbf{2.66 $\pm$ 0.24} && 
         \textbf{5.98 $\pm$ 0.35} & \textbf{6.29 $\pm$ 0.41} & 2.06 $\pm$ 0.35 && 
         5.57 $\pm$ 0.21 & 5.63 $\pm$ 0.38 & \textbf{1.27 $\pm$ 0.57} \\
         & PeRP &
         5.71 $\pm$ 0.61 & 6.14 $\pm$ 0.64 & 2.69 $\pm$ 0.28 && 
         4.63 $\pm$ 0.39 & 4.90 $\pm$ 0.45 & 1.92 $\pm$ 0.40 && 
         \textbf{6.41 $\pm$ 0.36} & \textbf{6.59 $\pm$ 0.43} & 1.69 $\pm$ 0.66 \\
         
         \midrule
         \multirow{4}{*}{-5.0}& PCP &  
         4.47 $\pm$ 0.38 & 4.91 $\pm$ 0.42 & 2.80 $\pm$ 0.24 && 
         4.58 $\pm$ 0.23 & 4.89 $\pm$ 0.29 & 2.08 $\pm$ 0.30 && 
         3.37 $\pm$ 0.31 & 3.67 $\pm$ 0.42 & 2.03 $\pm$ 0.52 \\ 
         & TA-RP &
         7.30 $\pm$ 0.14 & 7.62 $\pm$ 0.13 & 2.11 $\pm$ 0.18 && 
         5.21 $\pm$ 0.66 & 5.72 $\pm$ 0.69 & 3.31 $\pm$ 0.28 && 
         \textbf{7.98 $\pm$ 0.17} & \textbf{8.11 $\pm$ 0.09} & \textbf{1.37 $\pm$ 0.34} \\ 
         & RP &
         \textbf{7.60 $\pm$ 0.34} & \textbf{7.91 $\pm$ 0.26} & 2.08 $\pm$ 0.45 &&
         \textbf{7.56 $\pm$ 0.24} & \textbf{7.85 $\pm$ 0.21} & \textbf{1.96 $\pm$ 0.18} && 
         7.75 $\pm$ 0.33 & 7.90 $\pm$ 0.27 & 1.46 $\pm$ 0.36 \\ 
         & PeRP &
         7.20 $\pm$ 0.31 & 7.52 $\pm$ 0.32 & \textbf{2.07 $\pm$ 0.15} && 
         6.66 $\pm$ 0.48 & 7.02 $\pm$ 0.51 & 2.28 $\pm$ 0.47 && 
         7.54 $\pm$ 0.46 & 7.86 $\pm$ 0.40 & 2.12 $\pm$ 0.37 \\
         
         \midrule
         \multirow{4}{*}{-2.5}& PCP &  
         6.80 $\pm$ 0.33 & 7.15 $\pm$ 0.33 & 2.25 $\pm$ 0.14 && 
         7.04 $\pm$ 0.64 & 7.35 $\pm$ 0.65 & 2.03 $\pm$ 0.19 && 
         5.70 $\pm$ 0.23 & 5.76 $\pm$ 0.42 & \textbf{1.32 $\pm$ 0.65} \\ 
         & TA-RP &
         7.49 $\pm$ 0.17 & 7.91 $\pm$ 0.12 & 2.64 $\pm$ 0.32 && 
         6.48 $\pm$ 0.53 & 6.95 $\pm$ 0.52 & 3.00 $\pm$ 0.31 && 
         \textbf{7.64 $\pm$ 0.17} & \textbf{8.04 $\pm$ 0.11} & 2.56 $\pm$ 0.51 \\ 
         & RP &
         7.59 $\pm$ 0.26 & 8.01 $\pm$ 0.20 & 2.65 $\pm$ 0.40 && 
         7.43 $\pm$ 0.39 & 7.81 $\pm$ 0.35 & 2.45 $\pm$ 0.36 && 
         7.54 $\pm$ 0.34 & 7.95 $\pm$ 0.30 & 2.63 $\pm$ 0.41 \\
         & PeRP &
         \textbf{7.69 $\pm$ 0.19} & \textbf{8.03 $\pm$ 0.11} & \textbf{2.24 $\pm$ 0.53} && 
         \textbf{7.60 $\pm$ 0.64} & \textbf{7.88 $\pm$ 0.61} & \textbf{1.94 $\pm$ 0.34} && 
         7.54 $\pm$ 0.28 & 7.97 $\pm$ 0.21 & 2.76 $\pm$ 0.46 \\
         
         \midrule
         \multirow{4}{*}{0}& PCP &  
         7.45 $\pm$ 0.86 & 7.74 $\pm$ 0.72 & \textbf{2.12 $\pm$ 0.82} && 
         6.95 $\pm$ 1.35 & 7.26 $\pm$ 1.31 & \textbf{2.10 $\pm$ 0.39} && 
         \textbf{7.50 $\pm$ 1.19} & 7.65 $\pm$ 1.07 & \textbf{1.51 $\pm$ 0.56} \\ 
         & TA-RP &
         \textbf{7.63 $\pm$ 0.16} & \textbf{8.06 $\pm$ 0.10} & 2.71 $\pm$ 0.33 && 
         7.10 $\pm$ 0.57 & 7.45 $\pm$ 0.55 & 2.27 $\pm$ 0.34 && 
         7.23 $\pm$ 0.49 & 7.62 $\pm$ 0.53 & 2.50 $\pm$ 0.50 \\ 
         & RP &
         7.51 $\pm$ 0.34 & 7.97 $\pm$ 0.29 & 2.87 $\pm$ 0.47 && 
         \textbf{7.44 $\pm$ 0.17} & \textbf{7.91 $\pm$ 0.19} & 2.99 $\pm$ 0.29 && 
         7.47 $\pm$ 0.37 & 7.91 $\pm$ 0.36 & 2.83 $\pm$ 0.56 \\ 
         & PeRP &
         7.48 $\pm$ 0.33 & 7.93 $\pm$ 0.30 & 2.85 $\pm$ 0.35 && 
         7.22 $\pm$ 0.61 & 7.65 $\pm$ 0.65 & 2.67 $\pm$ 0.32 && 
         7.47 $\pm$ 0.31 & \textbf{7.93 $\pm$ 0.28} & 2.92 $\pm$ 0.50 \\
         
         \midrule
         \multirow{4}{*}{2.5}& PCP &  
         \textbf{7.61 $\pm$ 0.29} & 8.02 $\pm$ 0.23 & \textbf{2.59 $\pm$ 0.45} && 
         7.59 $\pm$ 0.15 & 8.01 $\pm$ 0.12 & 2.64 $\pm$ 0.25 && 
         \textbf{7.72 $\pm$ 0.13} & \textbf{8.11 $\pm$ 0.08} & \textbf{2.48 $\pm$ 0.31} \\ 
         & TA-RP &
         7.60 $\pm$ 0.24 & \textbf{8.04 $\pm$ 0.16} & 2.80 $\pm$ 0.53 && 
         \textbf{7.71 $\pm$ 0.27} & 8.01 $\pm$ 0.17 & \textbf{2.06 $\pm$ 0.51} && 
         7.69 $\pm$ 0.19 & 8.09 $\pm$ 0.13 & 2.55 $\pm$ 0.48 \\ 
         & RP &
         7.45 $\pm$ 0.21 & 7.95 $\pm$ 0.17 & 3.18 $\pm$ 0.31 && 
         7.60 $\pm$ 0.24 & \textbf{8.05 $\pm$ 0.17} & 2.85 $\pm$ 0.44 && 
         7.45 $\pm$ 0.18 & 7.95 $\pm$ 0.14 & 3.14 $\pm$ 0.32 \\ 
         & PeRP&
         7.53 $\pm$ 0.28 & 8.00 $\pm$ 0.19 & 3.03 $\pm$ 0.56 && 
         7.45 $\pm$ 0.19 & 7.93 $\pm$ 0.13 & 3.06 $\pm$ 0.43 && 
         7.47 $\pm$ 0.26 & 7.96 $\pm$ 0.20 & 3.13 $\pm$ 0.46 \\
         
         \midrule
         \multirow{4}{*}{5.0}& PCP &  
         7.50 $\pm$ 0.18 & 7.99 $\pm$ 0.14 & 3.10 $\pm$ 0.29 && 
         7.43 $\pm$ 0.15 & 7.92 $\pm$ 0.11 & 3.15 $\pm$ 0.33 && 
         \textbf{7.62 $\pm$ 0.24} & \textbf{8.07 $\pm$ 0.16} & 2.90 $\pm$ 0.52 \\ 
         & TA-RP &
         \textbf{7.57 $\pm$ 0.20} & \textbf{8.03 $\pm$ 0.14} & \textbf{2.94 $\pm$ 0.38} && 
         7.47 $\pm$ 0.18 & 7.91 $\pm$ 0.13 & \textbf{2.78 $\pm$ 0.32} && 
         7.54 $\pm$ 0.18 & 8.00 $\pm$ 0.14 & 2.94 $\pm$ 0.30 \\ 
         & RP &
         7.45 $\pm$ 0.17 & 7.95 $\pm$ 0.14 & 3.18 $\pm$ 0.24 && 
         \textbf{7.59 $\pm$ 0.25} & \textbf{8.05 $\pm$ 0.19} & 2.93 $\pm$ 0.45 && 
         7.60 $\pm$ 0.29 & 8.05 $\pm$ 0.20 & \textbf{2.88 $\pm$ 0.55} \\ 
         & PeRP &
         7.55 $\pm$ 0.24 & \textbf{8.03 $\pm$ 0.17} & 3.01 $\pm$ 0.45 && 
         7.55 $\pm$ 0.24 & 8.02 $\pm$ 0.18 & 2.98 $\pm$ 0.43 && 
         7.54 $\pm$ 0.22 & 8.02 $\pm$ 0.16 & 3.07 $\pm$ 0.40 \\

         \midrule
         \multirow{4}{*}{7.5}& PCP &  
         7.40 $\pm$ 0.15 & 7.92 $\pm$ 0.13 & 3.28 $\pm$ 0.22 && 
         7.52 $\pm$ 0.25 & 7.99 $\pm$ 0.18 & 3.01 $\pm$ 0.43 && 
         7.49 $\pm$ 0.26 & 7.98 $\pm$ 0.19 & 3.09 $\pm$ 0.47 \\ 
         & TA-RP &
         7.46 $\pm$ 0.14 & 7.96 $\pm$ 0.12 & 3.19 $\pm$ 0.17 && 
         \textbf{7.62 $\pm$ 0.23} & \textbf{8.07 $\pm$ 0.17} & \textbf{2.85 $\pm$ 0.43} && 
         \textbf{7.53 $\pm$ 0.20} & 8.00 $\pm$ 0.14 & \textbf{3.01 $\pm$ 0.43} \\ 
         & RP &
         7.43 $\pm$ 0.19 & 7.93 $\pm$ 0.14 & 3.19 $\pm$ 0.36 && 
         7.61 $\pm$ 0.21 & \textbf{8.07 $\pm$ 0.16} & 2.94 $\pm$ 0.36 && 
         7.51 $\pm$ 0.22 & 8.00 $\pm$ 0.17 & 3.11 $\pm$ 0.39 \\ 
         & PeRP &
         \textbf{7.52 $\pm$ 0.21} & \textbf{8.01 $\pm$ 0.16} & \textbf{3.08 $\pm$ 0.33} && 
         7.52 $\pm$ 0.23 & 7.99 $\pm$ 0.16 & 3.02 $\pm$ 0.44 && 
         \textbf{7.53 $\pm$ 0.27} & \textbf{8.01 $\pm$ 0.19} & 3.09 $\pm$ 0.53 \\
         
         \bottomrule
    \end{tabular}
    }
    \label{tab:quantitative-sim-driver-intent}
\end{table*}




\end{document}